\documentclass[10pt]{article}
\usepackage{ucs}
\usepackage[utf8x]{inputenc}
\usepackage[T1]{fontenc}
\usepackage{rev_sbrn}
\usepackage{times,amsfonts,enumerate,amssymb,amsmath,epsfig,epstopdf,subfigure,graphicx,rotating,multirow,bm,cite}
\usepackage[brazil]{babel}
\usepackage[ruled,vlined,linesnumbered]{algorithm2e}
\usepackage[a4paper, 
hmargin={2cm,1cm}, 
vmargin={2cm,2cm},
footskip=5mm]{geometry}

\begin{document}

\title{M{\'E}TODOS DE AGRUPAMENTOS EM DOIS EST{\'A}GIOS}
\author{Jefferson R. de Souza e Teresa B. Ludermir}
%\institute{Centro de Informática \\ Universidade Federal de Pernambuco, \email{tbl@cin.ufpe.br}}
%\and Name of Second Author \at Name, Address of Institute \email{name@email.address}}
\maketitle

\paragraph {{\bf Abstract}}This work investigates the use of two-stage clustering methods. Four techniques were proposed: SOMK, SOMAK, ASCAK and SOINAK. SOMK is composed of a SOM (Self-Organizing Maps) followed by the K-means algorithm, SOMAK is a combination of SOM followed by the Ant K-means (AK) algorithm, ASCAK is composed by the ASCA (Ant System-based Clustering Algorithm) and AK algorithms, SOINAK is composed by the Self-Organizing Incremental Neural Network (SOINN) and AK. SOINAK presented a better performance among the four proposed techniques when applied to pattern recognition problems.

\paragraph {{\bf Resumo:}} Este trabalho investiga a utilização de métodos de agrupamentos em dois estágios. Quatro técnicas foram propostas: SOMK, SOMAK,ASCAK e SOINAK. SOMK é composta de uma SOM (Mapas Auto-Organizáveis) seguido do algoritmo K-médias, SOMAK é uma combinação de SOM seguido do algoritmo Ant K-médias (AK), ASCAK é composto pelos algoritmos ASCA (Ant System-based Clustering Algorithm) e AK, SOINAK é composto pela Rede Neural Incremental Auto-Organizável (SOINN) e AK. SOINAK apresentou um melhor desempenho entre as quarto técnicas propostas quando aplicados nos problemas de reconhecimento de padrões.

\paragraph {{\bf Keywords:}} Sistemas Híbridos Inteligentes, Aprendizagem Não Supervisionada, K-médias, Self-Organizing Maps, Ant Colony Optimization, Ant System-based Clustering Algorithm, Self-Organizing Incremental Neural Network

% Page style "cbrna" defines the header for the first page
% The other pages should have an empty header
\thispagestyle{cbrna}
\pagestyle{cbrna}

% Set below the page number for the first page of your article
% by setting the page counter.
\setcounter{page}{1}
% Remember to edit the style file (.sty) such that the header in
% the first page shows the correct first and last page numbers of
% the article

\Section{INTRODU\c{C}\~{A}O}
\label{sec:intro}
Clusterização é uma técnica de aprendizado não-supervisonado que faz o agrupamento automático de instâncias similares, sem a necessidade de que os dados sejam rotulados. Ou seja, um algoritmo que clusteriza dados classifica eles em conjuntos de dados que ‘se assemelham’ de alguma forma - independentemente de classes predefinidas. Os grupos gerados por essa classificação são chamados clusters \cite{nath:Jain1999} e \cite{nath:Braga2000}. Padrões pertencentes a um dado \emph{cluster} devem ser mais similares entre si do que em relação a padrões pertencentes a outros clusters \cite{nath:Moscato2002}.

A análise de agrupamento de dados é um importante domínio de pesquisa de mi\-ne\-ra\-ção de dados. Esta análise tem aplicação em muitos domínios, tais como: Negócio, Medicina, Geografia e Processamento de Imagens. Além disso, vem sendo usada em vários campos, como: Estatística, Reconhecimento de Padrões e Aprendizado de Máquina. 

Sistemas Híbridos Inteligentes (SHIs) \cite{nath:Rezende2002}, \cite{GOMES-2013}, \cite{Aida-ijcnn-2009} e \cite{MINKU-2008} são modelos que resultam da combinação de duas ou mais técnicas com\-pu\-ta\-cio\-nais distintas para solucionar um dado problema. O uso destes sistemas é proposto por combinar as vantagens de suas técnicas, e superar as limitações que cada uma apresenta individualmente na resolução do problema de interesse. Estes sistemas podem nos levar a soluções mais robustas e eficientes, quando aplicados cor\-re\-ta\-men\-te. Acredita-se, que um modelo de computação inteligente tem grandes chances de se desenvolver se unir os seus potenciais com os diferentes modelos da Inteligência Artificial. Vale salientar, que a utilização dos SHIs não nos impulsiona a uma melhora do desempenho do sistema como um todo, muitas vezes os resultados obtidos pelos SHIs são inferiores ao desempenho das técnicas que o compõem \cite{nath:Braga2000}.

Neste trabalho, diversas técnicas foram investigadas de modo a contribuir para o a\-pri\-mo\-ra\-men\-to do sistema de descoberta de padrões, visando utilizá-las para solucionar o problema de agrupamento. A primeira técnica é a SOMK, composta da rede SOM seguida do algoritmo K-médias, onde SOMK é útil para diminuição de ruídos \emph{outliers} nos dados \cite{nath:Vesanto2000}. SOMK se baseia em dois estágios, onde o primeiro é aplicado para o pré-processamento dos dados e o segundo para agrupar os dados finais. A segunda técnica observada é o método de agrupamento SOMAK que utiliza uma rede SOM com o procedimento AK. SOMAK tem algumas vantagens tais como: a redução do tempo computacional, a redução da quan\-ti\-da\-de de a\-gru\-pa\-men\-tos e a redução de ruídos, os quais são observações num conjunto de dados, que é suficientemente dissimilar do restante dos dados, ou seja, estas observações são designadas por observações anormais, con\-ta\-mi\-nan\-tes, estranhas e extremas) \cite{nath:Souza2009a} \cite{nath:Souza2009b}. A terceira técnica é a ASCAK que resulta da combinação dos algoritmos ASCA (do inglês, \emph{Ant System-based Clustering Algorithm}) e AK. ASCAK apresentou melhores resultados quando comparado ao sistema híbrido SOMAK. E por fim, a quarta técnica é a SOINAK, composta da rede SOINN \cite{nath:Furao2006}, que aplica aprendizagem não-supervisionada de forma incremental seguida do AK.

SOMAK, ASCAK e SOINAK utilizam SOM, ASCA e SOINN respectivamente como métodos de primeiro estágio sobre os dados de entrada, ao invés de realizar o agrupamento dos dados diretamente. Primeiro, um conjunto grande de protótipos é formado pela rede SOM, ASCA e SOINN. Na segunda etapa, os protótipos são interpretados e então combinados para formar os agrupamentos finais. A segunda etapa do sistema híbrido proposto é o algoritmo AK, que se baseia na abordagem meta-heurística, proposta para solucionar problemas de difícil otimização combinatória, denominada Otimização baseada em Colônia de Formigas ou \emph{Ant Colony Optimization} (ACO) \cite{nath:Dorigo2000}. ACO é um método de otimização bio-ins\-pi\-ra\-do que toma como referência o comportamento das formigas, de forma a identificar caminhos mais curtos entre o ninho e a fonte de alimento.

Deve-se ressaltar que o objetivo neste trabalho não é o de encontrar um a\-gru\-pa\-men\-to ótimo para os dados, mas obter uma visão sobre a estrutura de agrupamento dos dados usando SOMAK, ASCAK e SOINAK, com a finalidade de reduzir o tempo computacional e melhorar o desempenho quando comparado com outras técnicas de agrupamento. As vantagens do SOMAK são a redução do tempo computacional, a redução da quantidade de a\-gru\-pa\-men\-tos e a redução de ruídos. A vantagem do método de agrupamento ASCAK é a redução de ruídos. A vantagem principal do SOINAK quando comparada com os modelos SOMK, SOMAK e ASCAK é o fato que não é necessário saber \emph{a priori} o número de agrupamentos como no algoritmo K-médias (ou AK). Outro benefício importante é que não é necessário informar os centróides iniciais com SOINAK.

Este artigo está dividido em seis seções. A seção 2 mostra as técnicas de aprendizado não-supervisionado e sistemas híbridos. Na seção 3 são descritos os métodos propostos. Na seção 4 são apresentados os materiais e métodos usados neste trabalho. A seção 5 apresenta os resultados obtidos de cada um dos métodos de agrupamentos em problemas de reconhecimento de padrões. As considerações finais são apresentadas na seção 6.

\section{Aprendizado Não-Supervisionado}
\label{sec:aprendizadoNS}
No aprendizado não-supervisionado não há um supervisor externo para acompanhar o processo de aprendizado. Neste esquema de treinamento, somente os padrões de entrada estão disponíveis, ao contrário do aprendizado su\-per\-vi\-sio\-na\-do, cujo conjunto de treinamento contém pares de entrada e saída. É necessário que exista uma regularidade e redundância nos padrões de entrada, pois são características fundamentais para haver aprendizado não-supervisionado \cite{nath:Braga2000}. Este tipo de aprendizado é aplicado aos problemas que se preocupam com a des\-co\-ber\-ta de agrupamentos (grupos). Modelos bastante conhecidos e usados são: K-médias \cite{nath:Mitchell1997}, SOM de Kohonen \cite{nath:Kohonen1998} e SOINN de Furao e Hasegawa \cite{nath:Furao2006}.

\subsection{K-médias}
O K-médias é um dos mais simples algoritmos de aprendizagem não-supervisionada usados para solucionar o problema de agrupamento. O objetivo é dividir os dados dentro de \emph{k} grupos fixados \emph{a priori}. O al\-go\-ri\-tmo consiste em dois estágios: um inicial e um iterativo. O estágio inicial envolve a definição dos \emph{k} centros ou centróides, um para cada agrupamento. Muitas vezes, os centróides são escolhidos aleatoriamente. A escolha de diferentes centróides iniciais pode levar à resultados finais diferentes. O segundo estágio do algoritmo K-médias consiste em repetir a diferença entre o padrão de entrada e o centróide mais próximo, com isso calculando os \emph{k} novos centróides \cite{nath:Mitchell1997}. Esta iteração pára quando certo critério é estabelecido; por exemplo, número de iterações ou não haver mudança na posição dos centróides. Suponhamos que em um dado conjunto \emph{N} (amostras de dados), que\-re\-mos classificar os dados dentro de \emph{k} (grupos), o algoritmo tende a minimizar uma função de erro, como, por exemplo, o erro médio quadrático definido na Eq1:

\begin{center}
    \begin{equation}
        E = \sum_{j=1}^{k} \sum_{i=1}^{N} || x_{\imath} - c_{\jmath} ||^{2}
    \end{equation}
\end{center}

onde \textit{k} representa o número de agrupamentos, \textit{N} o número de amostras, \textit{x} a entrada de cada amostra e $ c_{\jmath} $ é o centro de $\jmath$ agrupamentos. O Algoritmo ~\ref{Alg:Kmedias}, usado para o treinamento do algoritmo K-médias, é resumido a seguir.

\begin{algorithm}
Definição dos \emph{k} centros aleatoriamente \\
\While{Não houver mudança nos centros}{
    \For{Cada padrão $x_{\imath}$ de treinamento}{
        Atribuir $x_{\imath}$ ao grupo associado com o centro $c_{\jmath}$ mais próximo \\
        Calcular um novo centro para cada grupo \\
        \For {cada grupo $c_{\jmath}$}{
            Calcular $c_{\jmath} = \sum_{j=1}^{k} \frac{|| x_{\imath} - c_{\jmath} ||}{N}$ \\
        }
    }
}
\caption{Algoritmo K-médias \cite{nath:Mitchell1997}}
\label{Alg:Kmedias}
\end{algorithm}

\subsection{SOM}
A rede neural auto-organizável (SOM, do inglês, \emph{Self-Organizing Maps}) \cite{nath:Kohonen1998} consiste em uma grade regular, unidades de processamento sobre o mapa e normalmente possui duas dimensões (2-D). Cada unidade de processamento \emph{i} é representada por um vetor de protótipos (pesos da conexão) $ m_{i} = [m_{i1},..., m_{id}] $, onde \emph{d} é a entrada da dimensão do vetor. As unidades estão conectadas a seus adjacentes por uma relação de vizinhança. O número das unidades do mapa varia de algumas dezenas até milhares, o qual determina a acurácia e a capacidade de generalização da rede SOM. Durante o treinamento, SOM forma uma rede elástica que se dobra, composta pelos dados de entrada. Os dados situados próximos uns dos outros no espaço de entrada são mapeados próximos das unidades do mapa. Assim, SOM pode ser interpretada como uma topologia, preservando o mapeamento do espaço de entrada sobre a grade 2-D das unidades do mapa.

SOM é treinada iterativamente, ou seja, em cada etapa de treinamento um vetor de amostra (padrões) \emph{x} é escolhido aleatoriamente de um conjunto de dados de entrada. As distâncias entre \emph{x} e todos os vetores de protótipos são calculados. A melhor unidade de processamento (\emph{Best Matching Unit} (BMU)) é denotada por \emph{b} que é a unidade do mapa com o protótipo mais próximo de \emph{x}. A seguir serão mostradas as variáveis fundamentais para o entendimento do Algoritmo ~\ref{Alg:SOM}, são elas: \emph{t}: tempo, $\alpha(t)$: coeficiente de adaptação (taxa de aprendizado), $h_{bi}(t)$: vizinhança de \emph{kernel} centrada sobre a unidade vencedora (neurônio ven\-ce\-dor), $r_{b}$ e $r_{i}$: posições dos neurônios \emph{b} e \emph{i} sobre a grade da rede neural SOM, \emph{N}: número de padrões de treinamento e \emph{M}: número das unidades do mapa. O Algoritmo ~\ref{Alg:SOM}, utilizado para o treinamento da rede neural SOM, é resumido a seguir.

\begin{algorithm}
Inicialização dos pesos e parâmetros \\
\While{Mapa de características não modificar}{
    \For{Cada padrão de treinamento}{
        Definir neurônio vencedor: $\emph{b} = arg  min_{i}{||x - m_{i}||}$\\
        Atualizar os pesos deste neurônio e de seus vizinhos \\
        $m_{i}(t+1) = m_{i}(t) + \alpha(t)h_{bi}(t)[x - m_{i}(t)]$ e $h_{bi}(t) = exp(-\frac{||r_{b} - r_{i}||^2}{2\sigma^2(t)})$ \\
        \If{O número do ciclo for mútliplo de N}{reduzir a taxa de aprendizado}{}
    }
}
\caption{Algoritmo SOM \cite{nath:Kohonen1998}}
\label{Alg:SOM}
\end{algorithm}

No caso de dados discreto e vizinhança de \emph{kernel} fixa, a função de erro da rede SOM pode ser:

\vspace{-25pt}
\begin{center}
    \begin{equation}
        E = \sum_{i=1}^{N}\sum_{j=1}^{M} h_{bj}||x_{i} - m_{j}||^2
    \end{equation}
\end{center}

SOM pode ser aplicado em um grande conjunto de dados. A complexidade computacional cresce linearmente com o número de amostras de dados, e não requer uma grande quantidade de memória (basicamente apenas os vetores de protótipos e o vetor de treinamento atual), e pode ser implementada tanto em uma forma de aprendizagem neural ou on-line, bem como em paralelo \cite{nath:Lawrence1999}. Por outro lado, a com\-ple\-xi\-da\-de cresce quadraticamente com o número de unidades de processamento do mapa. Assim, mapas de treinamento muito grandes consomem muito tempo, embora o processo possa ser acelerado com técnicas especiais \cite{nath:Kohonen2000} e \cite{nath:Koikkalainen1995}.

\subsection{SOINN}
Furao e Hasegawa \cite{nath:Furao2006} propuseram uma rede neural incremental auto-or\-ga\-ni\-zá\-vel (SOINN, do inglês, \emph{Self-Organizing Incremental Neural Network}), visando realizar a aprendizagem não-su\-per\-vi\-sio\-na\-da de dados. SOINN pode ser usado para solucionar problemas de aprendizagem não supervisionada, gerar grupos de dados correlacionados e realizar esse aprendizado de forma incremental. Assim, a técnica intenciona alcançar os seguintes objetivos: não necessitar de conhecimento, \emph{a priori}, dos dados (por exemplo, o número de grupos), ou seja, ser minimamente dependente da entrada paramétrica dos dados pelo usuário e separar grupos com interseções de baixa densidade e reportar re\-pre\-sen\-tan\-tes adequados para cada grupo gerado.

\begin{figure}[h!]
\begin{center}
\includegraphics[height=6cm]{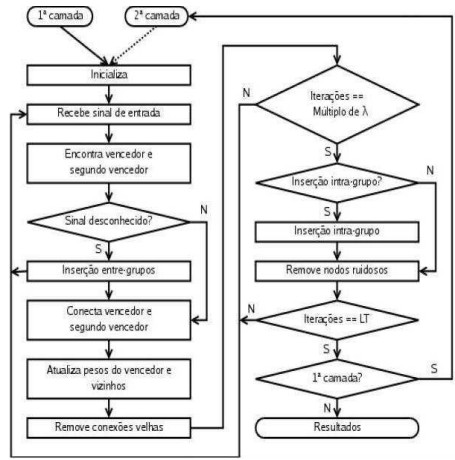}
\end{center}
\vspace{-20pt}
\caption{Fluxo de execução do algoritmo SOINN \cite{nath:Furao2006}}
\label{fig:ProcedimentoSOINN}
\end{figure}

O SOINN adota uma estrutura de duas camadas: a primeira aprende a distribuição da densidade dos dados de entrada e a segunda realiza um refinamento do conhecimento adquirido pela primeira, reportando os grupos formados pelas entradas submetidas até então, com uma quantidade menor de nodos (ou neurônios) que a reportada pela primeira camada para representar a topologia dos dados. O método de aprendizagem usado para a segunda camada é o mesmo da primeira, recebendo como entrada o conjunto de dados (pesos dos nodos) aprendidos na primeira camada. A Figura~\ref{fig:ProcedimentoSOINN} apresenta um diagrama de fluxo do algoritmo.

Ao receber uma nova entrada, são encontrados dois nodos chamados de primeiro e segundo vencedor. Em seguida, o algoritmo verifica se esta entrada pertence ou não a um dos grupos (primeiro ou segundo vencedor), de acordo com a proximidade deles, mediante a utilização de um limiar adaptativo, que é ajustado dinamicamente na primeira camada para cada nodo.

Caso a entrada não se encontre dentro dos limites do primeiro e segundo vencedores, ela é adicionada à rede como um novo nodo, e o algoritmo segue para processar uma nova entrada. Esta inserção é chamada de inserção entre-grupos. Por outro lado, se a entrada é considerada como pertencente a um dos dois grupos e uma conexão entre o primeiro e o segundo ven\-ce\-dor não existir, ela é criada, e, tal conexão possui sua variável \emph{idade} zerada. Depois, os pesos do nodo vencedor são ajustados usando uma taxa de aprendizagem adaptativa e os de seus vizinhos são modificados, usando uma segunda taxa de aprendizagem. Há diversas variáveis, como: densidade local acumulada, útil para realizar a contagem de vitórias do nodo e o Erro Local Acumulado (E). Além disso, as idades de todas as conexões que herdam do nodo vencedor devem ser incrementadas.

Todas as conexões com \emph{idade} superior a um valor constante predefinido, \emph{$age_{dead}$}, são removidas. Após um múltiplo de $\lambda$ (lambda) iterações de aprendizagem, sendo $\lambda$ uma constante predefinida pelo usuário, SOINN realiza uma tentativa de inserção de um novo nodo entre o maior E e seu vizinho com maior E. Essa inserção, chamada de inserção intra-grupo, é cancelada se não conseguir diminuir o erro local, que é calculado para cada nodo como a razão entre o E e sua respectiva Densidade Local Acumulada (M). Depois, SOINN encontra todos os nodos com nenhuma conexão e os apaga. Para nodos com apenas uma conexão, o nodo é apagado se a M for menor que a média da densidade acumulada de todos os nodos da rede. Este passo ocasiona a remoção de prováveis ruídos e áreas de baixa densidade de dados.

Após \emph{LT} iterações, os pesos dos nodos são submetidos como entrada para a segunda camada, que, realiza um processo de aprendizagem da mesma forma da primeira camada, gerando, ao fim, grupos mais coesos. Portanto, reporta os grupos gerados e seus representantes adequados. Tais grupos são formados por nodos que estão conectados por meio de um caminho de conexões. A única distinção entre o processo de aprendizagem das camadas é o limiar adotado, que na primeira camada é adaptativo e na segunda é constante. A segunda camada da rede SOINN proporciona um refinamento do resultado obtido pela primeira, eliminando dados ruidosos e irrelevantes, mantendo apenas os que são representativos para cada grupo.

\subsubsection{Implementação do Algoritmo SOINN}

Um roteiro do funcionamento do algoritmo SOINN é descrito a seguir. A diferença entre as duas camadas é que o conjunto de dados de entrada da segunda camada são os nodos resultantes da primeira camada. Além disso, ao invés do limiar dinâmico da primeira camada, é usado um limiar estático na segunda. Algumas notações são usadas durante a descrição do algoritmo: A: conjunto de nodos, $N_{A}$: número de nodos de A, C: conjunto de conexões, $N_{C}$: número de conexões em C, $W_{i}$: vetor de pesos n-dimensional do nodo \emph{i}, $E_{i}$: erro local acumulado do nodo \emph{i} (atualizado quando o nodo \emph{i} é o vencedor), $M_{i}$: densidade local acumulada do nodo \emph{i} (representa o número de vezes que o nodo \emph{i} foi vencedor), $R_{i}$: raio de erro do nodo \emph{i} (definido como a média do erro acumulado, $\frac{E_{i}}{M_{i}}$, e atualizada na inserção), $G_{i}$: rótulo do grupo (usada para verificar qual grupo o nodo \emph{i} pertence), Q: número de grupos, $T_{i}$: limiar de similaridade, $N_{i}$: conjunto de vizinhos topológicos diretos do nodo \emph{i}, $L_{i}$: número de vizinhos topológicos do nodo \emph{i}, $\emph{age}_{(i,j)}$: idade da conexão entre o nodo \emph{i} e o nodo \emph{j} e \emph{path}: dado uma série de nodos $x_{i} \in A$, \emph{i} = 1, ..., n, fazendo (\emph{i}, x1), (x1, x2),..., (xn-1, xn), (xn, \emph{j}) C. Nesse caso, existe um caminho entre o nodo \emph{i} e \emph{j}. O Algoritmo ~\ref{Alg:SOINN}, usado para o treinamento de SOINN, é resumido a seguir.

\begin{algorithm}
contadorCamadas = 1 \\
\While{(contadorCamadas $\le 2$)}{
    \eIf{(contadorCamadas == 1) ou (contadorCamadas == 2)}{
        Inicialize o conjunto A = {$c_{1}$, $c_{2}$} com pesos aleatórios do padrão de entrada e o conjunto de conexões C = $\emptyset$ \\
        \For{Cada padrão $\xi_{i}$ de treinamento}{
            Procure A determinando os nodos: $s_{1} = arg  min_{c \in A}\parallel \xi - A_{c}\parallel$ e $s_{2} = arg  min_{c \in A} \setminus s_{1} \parallel \xi - A_{c}\parallel$ \\
            \eIf{($||\xi - s_{1}|| > T_{s_{1}}$) ou ($||\xi - s_{2}|| > T_{s_{2}}$)}{A = {$\xi$}}{
                $C = C \cup (s_{1}, s_{2})$, $\emph{age}_{(s_{1}, \emph{i})} = \emph{age}_{(s_{1}, \emph{i})} + 1 (\forall\emph{i} \in \emph{N}_{s_{1}})$, $\emph{E}_{s_{1}} = \emph{E}_{s_{1}} + \parallel \xi - \emph{W}_{s_{1}} \parallel$ e $M_{s_{1}} = M_{s_{1}} + 1$ \\ $\Delta\emph{W}_{s_{1}} = \varepsilon_{1}(\emph{t})(\xi - \emph{W}_{s_{1}})$, $\Delta\emph{W}_{i} = \varepsilon_{2}(\emph{t})(\xi - \emph{W}_{i}) (\forall_{i} \in \emph{N}_{s_{1}})$ e remove conexões com \emph{idade} $ > \emph{age}_{dead}$ \\
                \If{($\xi_{i}$ for múltiplo de $\lambda$)}{
                    q = arg $max_{c \in A}E_{c}$, f = arg $max_{c \in N_{q}}E_{c}$, A = A $\cup r$, $W_{r} = (W_{q} + W_{f})/2.0$ e $E_{r} = \alpha_{1}(E_{q} + E_{f})$ \\ 
                    $M_{r} = \alpha_{2}(M_{q} + M_{f})$, $R_{r} = \alpha_{3}(R_{q} + R_{f})$, $E_{q} = \beta E_{q'}$, $E_{f} = \beta E_{f}$, $M_{q} = \gamma M_{q'}$ e $M_{f} = \gamma M_{f}$ \\
                    \If{$R_{r} > R_{i}$}{C = C$\cup (r,q)(r,f)$ e $C = C \setminus (q,f)$ \\}{}
                    \For{Todos os nodos em A}{\If{Nodo possuir um vizinho e $M_{r} <$ limiar adaptativo}{A = nodo}{}}
                }{}
            }
        }
    }{
        padrões de entrada = nodos
    }
    contadorCamadas = contadorCamadas + 1
}
%Inicializa todos os nodos como não classificados \\
\While(\tcp*[h]{Antes, inicialize todos os nodos como NC}){Todos os nodos (nós) estiverem classificados}{
    Escolha um nodo aleatório não classificado \emph{i} em A, marque nodo \emph{i} como classificado e rotule-o com $G_{i}$ \\
    Procure A nós Não Classificados (NC) ao nó \emph{i} por 1 caminho, marque nós classificados e rotule-os com classe do nó \emph{i} \\
}
\caption{Algoritmo SOINN \cite{nath:Furao2006}}
\label{Alg:SOINN}
\end{algorithm}

\subsubsection{Determinação dos Parâmetros}
%Os parâmetros utilizados no algoritmo foram determinados como segue.

\begin{enumerate}
  \item \textbf{Limiares de similaridade}

    Os limiares determinam a inserção ou não de um novo nodo na rede. Um limiar ótimo, é maior que a distância intra-gru\-po e menor que a distância entre-grupos. A primeira camada adota um limiar adaptativo, se o nodo possuir vizinhos topológicos, seu limiar $T_{i}$ é a maior distância entre ele e seus vizinhos $(\emph{T}_{i}=max_{c\in\emph{N}_{i}}\parallel\emph{W}_{i}-\emph{W}_{c}\parallel)$;  caso o nodo não possua vizinhos seu limiar $T_{i}$ é a distância mínima entre ele e todos os nodos da rede $(\emph{T}_{i}=min_{c\in\emph{A}}$ \ ${i}\parallel\emph{W}_{i}-\emph{W}_{c}\parallel)$. A segunda camada usa um limiar constante $T_{c}$, que é a menor distância entre os grupos (e.g. maior distância entre elementos do 1º e 2º grupo), maior que a Média das Distâncias Intra-grupo (MDI). MDI é a média das distâncias entre os nodos de todas as conexões. 

  \item \textbf{Taxa de aprendizagem adaptativa}

    SOINN usa uma Taxa de Aprendizagem Adaptativa (TAA), proporcionando os nodos uma parcela maior dos dados e menos adaptados para representar novos padrões de entrada; assim, aumentando a estabilidade da rede. Isto proporciona a capacidade de aprendizado de SOINN. SOINN usa uma TAA por nodo que decai com o tempo (\emph{t}), considerando M nodo, que indica quantas vezes o nodo foi vencedor, assim \emph{t} = $M_{i}$. $\varepsilon_{1}(t) = \frac{1}{t}$ representa a Função da Taxa de Aprendizagem (FTA) para o nodo vencedor e $\varepsilon_{2}(t) = \frac{1}{100*t}$ a FTA para os nodos que são vizinhos diretos do nodo vencedor.

  \item \textbf{Fatores de decaimento de E, M e R}

    Quando realiza uma inserção intra-grupo usa-se $\alpha1$, $\alpha2$ e $\alpha3$ para estabelecer os valores de E, M e o raio de erro herdado (R) do novo nodo r, como parcelas dos valores dos nodos q e f. E e M decrescem nos nodos q e f pelas frações de $\beta$ e $\gamma$, visto que estes valores são divididos com o novo nodo \emph{r}. Tais valores foram determinados em \cite{nath:Furao2006} através de uma análise de divisão de regiões de Voronoi: $\alpha1 = 1/6$, $\alpha2 = 1/4$, $\alpha3 = 1/4$, $\beta = 2/3$ e $\gamma = 3/4$. %Estes valores não devem ser considerados como os melhores ajustes para todos os casos, porém podem ser usados em geral.
\end{enumerate}

\subsection{Colônia de Formigas}
A Otimização Baseada em Colônia de Formigas ou Ant Colony Optimization (ACO) foi proposta  Dorigo \cite{nath:Dorigo2000} e tem sido uma das técnicas bem sucedidas em Sistemas Inteligentes de Enxames \cite{nath:Shelokar2004}. Estes sistemas são inspirados a partir da co\-la\-bo\-ra\-ção do comportamento social dos animais, como pássaros, peixes e formigas, e sua incrível formação de bandos, enxames e colônias. 

O Algoritmo ~\ref{Alg:ACO} apresenta o funcionamento do ACO. Os feromônios ($\tau_{ij}$) são inicializados por um valor constante ($\tau_{0}$), ou seja, valores aleatórios dentro de um intervalo [0, $\tau_{0}$]. Pode-se observar, que $\emph{n}_{k}$ representa o número de formigas, $x^{k}(\emph{t})$ denota uma solução no tempo \emph{t}, \emph{f}($x^{k}(\emph{t})$) expressa a qualidade da solução e $\eta = \frac{1}{d_{\imath\jmath}}$, onde $d_{\imath\jmath}$ é a distância entre os nós $\imath$ e $\jmath$. O conjunto, $N^{k}_{i}$, define o conjunto de nós viáveis para \emph{k} formiga quando localizada pelo nó $\imath$. Os demais parâmetros, podem ser encontrados na seção do algoritmo \emph{Ant} K-médias por ser um algoritmo de agrupamento baseado nas colônias de formigas.

\begin{algorithm}
Estabeleça \emph{t} = 0, inicialize os parâmetros $\alpha$, $\beta$, $\rho$, Q, $\emph{n}_{k}$ e $\tau_{0}$ e coloque todas as formigas, \emph{k} = 1, ..., $\emph{n}_{k}$\\
\For{Cada \emph{link} ($\imath$,$\jmath$)}{
    $\tau_{ij}(\emph{t}) = \emph{U}(0, \tau_{0})$\\
}
\Repeat{Condição de parada for verdadeira}{
    \For{Cada formiga \emph{k} = 1, ..., $\emph{n}_{k}$}{
        $x^{k}(\emph{t}) = 0$\\
        \Repeat{O caminho completo estiver construído}{
            A partir do nó atual $\imath$, selecione o próximo nó $\jmath$ com probabilidade definida na equação abaixo\\
            $
            P_{\imath\jmath}^{k}(\emph{t}) = \left\{ \begin{array}{rl}
              \dfrac{\tau_{\imath\jmath}^{\alpha}(\emph{t}) \eta_{\imath\jmath}^{\beta}(\emph{t})} {\sum_{u}^{N^{k}_{i}(\emph{t})} \tau_{\imath u}^{\alpha}(\emph{t}) \eta_{\imath u}^{\beta}(\emph{t})} &\mbox{ if $\jmath \in N^{k}_{i}(\emph{t})$}\\
              0 &\mbox{caso contrário}
                   \end{array} \right.
            $\\

            $x^{k}(\emph{t}) = x^{k}(\emph{t}) U (\imath,\jmath)$\\
        }
        Calcular \emph{f}($x^{k}(\emph{t})$)\\
    }
    \For{Cada \emph{link} ($\imath$,$\jmath$)}{
        Aplicar a evaporação usando a equação: $\tau_{\imath\jmath}(\emph{t}) \leftarrow (1-\rho)*\tau_{\imath\jmath}(\emph{t})$\\
        Calcular $\Delta\tau_{\imath\jmath}(\emph{t})$ usando a equação: $\Delta\tau_{\imath\jmath}(\emph{t}) = \sum_{k=1}^{\emph{n}_{k}} \Delta\tau_{\imath\jmath}^{k}(\emph{t})$\\
        Onde, $ \Delta\tau_{\imath\jmath}^{k}(\emph{t}) = \left\{ \begin{array}{rl}
         \frac{Q}{\emph{f}(x^{k}(\emph{t}))} &\mbox{ if \emph{link} ($\imath$,$\jmath$) ocorrer no caminho $x^{k}(\emph{t})$} \\
          0 &\mbox{caso contrário}
               \end{array} \right.
        $\\
        Atualizar o feromônio usando a equação: $\tau_{\imath\jmath}(\emph{t} + 1) = \tau_{\imath\jmath}(\emph{t}) + \sum_{k=1}^{\emph{n}_{k}} \Delta\tau_{\imath\jmath}^{k}(\emph{t})$
    }
    \For{Cada \emph{link} ($\imath$,$\jmath$)}{
        $\tau_{\imath\jmath}(\emph{t} + 1) = \tau_{\imath\jmath}(\emph{t})$\\
    }
    \emph{t} = \emph{t} + 1\\
}
Retorna $x^{k}(\emph{t}) : \emph{f}(x^{k}(\emph{t})) = min_{k' = 1, ..., \emph{n}_{k}} \emph{f}(x^{k'}(\emph{t})$)\\
\caption{Algoritmo ACO \cite{nath:Dorigo2000}}
\label{Alg:ACO}
\end{algorithm}

As formigas se co\-mu\-ni\-cam entre si apenas de forma indireta através de seu ambiente, pela substância chamada feromônio, que são substâncias excretadas por organismos vivos e detectadas por outros indivíduos da mesma espécie, produzindo mudanças de comportamento específicos. Os feromônios atuam na comunicação intra-específica, ou seja, entre membros de uma mesma espécie. Assim, os caminhos com maiores níveis de feromônio serão mais prováveis de serem escolhidos e, portanto, reforçados, enquanto que a intensidade desta substância sobre os caminhos que não são escolhidos é reduzida pela e\-va\-po\-ra\-ção. Esta forma de co\-mu\-ni\-ca\-ção indireta é conhecida, e prevê à colônia de formiga a capacidade de encontrar o menor caminho ou percurso \cite{nath:Martens2007}.

ACO tem sido aplicado com sucesso em muitas situações, como: problema do caixeiro viajante, problemas de horário, dentre outros \cite{nath:Niknam2008}. Yuqing et al. \cite{nath:Yuqing2003} propuseram um novo algoritmo K-médias baseado na densidade e nas colônias de formigas. Este algoritmo solucionou o problema do mínimo local e a manipulação dos parâmetros iniciais sensíveis do K-médias. 

Existe uma variedade de problemas de otimização que são solucionados por outro tipo de técnica inteligente baseada em enxames, a qual é definida pelo algoritmo de Otimização baseado em Partículas de Enxames (PSO, do inglês, \emph{Particle Swarm Optimization}) \cite{nath:Saatchi2005}. PSO foi utilizado para treinamento de redes \emph{Multi-layer Perceptrons} \cite{Marcio-iconip2006} e para otimização de redes \emph{Multi-layer Perceptrons} com \emph{weight decay} \cite{Marcio-his2006}. Alguns outros pesquisadores propuseram um agrupamento de genes usando a rede SOM e PSO \cite{nath:Xiao2003}. Nesta combinação, foi desenvolvida uma abordagem de agrupamento híbrida. No algoritmo proposto por Xiao et al. \cite{nath:Xiao2003}, a taxa de convergência é melhorada pela adição de um fator de consciência para a rede SOM. A robustez do resultado é medida com a utilização de uma técnica de reamostragem. O método proposto SOM-PSO descrito por Xiao et al. preserva a estrutura de topologia da rede SOM e realiza tentativas de gerar um resultado de agrupamento mais compacto que a rede de Kohonen \cite{nath:Kohonen1998}. PSO foi utilizado para otimizar as topologias de  \emph{Extreme Learning Machines} \cite{Elliackin-sbrn-2012} com sucesso. Vários topologias foram analizadas no estudo. Chu et al. \cite{nath:Chu2004} estenderam o algoritmo ACO para o processo de agrupamento de dados para solucionar alguns problemas nos métodos de agrupamento con\-ven\-cio\-nais, ou seja, formas arbitrárias, agrupamentos com \textit{outliers}, entre outros. O \textit{Quadratic Distance Me\-tric (QDM)} permite limitar a adição de feromônio e escolher um limite es\-tra\-té\-gi\-co para melhorar os agrupamentos de dados.

\subsection{Sistemas Híbridos}
A motivação do desenvolvimento de Sistemas Híbridos (SHs) é que uma única técnica, pode não ser capaz de, sozinha, solucionar um dado problema. Assim, a combinação de duas ou mais técnicas podem conduzir a uma solução robusta e eficiente \cite{nath:Braga2000}. Neste artigo são descritos alguns SHs que se assemelham com os sistemas propostos nesta pesquisa.

Vesanto e Alhoniemi propuseram uma técnica de agrupamento que combina SOM \cite{nath:Kohonen1998} e K-médias \cite{nath:Mitchell1997}, os autores mostraram em seu artigo di\-fe\-ren\-tes abordagens para o agrupamento de dados usando SOM. O método proposto \cite{nath:Vesanto2000} é formado por dois estágios: o primeiro usa SOM para produzir os protótipos iniciais, os quais são posteriormente a\-gru\-pa\-dos, no segundo estágio, pelo K-médias. Portanto, os resultados observados neste artigo comprovam que o agrupamento usando SOM como uma etapa intermediária foi computacionalmente eficaz, ou seja, apresentou um bom de\-sem\-pe\-nho quando comparado com os resultados obtidos diretamente a partir dos dados (e.g. SOM e K-médias de forma isolada). O uso do SOM antes do K-médias minimizam as dificuldades do K-médias, que são: escolha inicial dos centróides, quantidade de centróides a serem informados, entre outros.

Em virtude dos problemas apresentados por Vesanto e Alhoniemi \cite{nath:Vesanto2000} e com a pers\-pe\-c\-ti\-va de uma melhora crescente no sistema híbrido para o reconhecimento de padrões, Kuo et al. propuseram um novo método de agrupamento, que se baseia nas colônias de formigas, conhecido na literatura como AK para a análise de agrupamento de dados \cite{nath:Kuo2005}. AK modifica K-médias localizando os objetos e em seguida agrupando-os de acordo com uma probabilidade. Esta probabilidade é atualizada a partir de um determinado feromônio conforme o Total de Variância dentro do Agrupamento (TWCV, do inglês, \emph{Total Within Cluster Variance}). Os resultados alcançados em \cite{nath:Kuo2005}, mostram que AK é melhor em sua eficiência e desempenho do que outros dois métodos: SOMK e a combinação da rede SOM seguida do algoritmo K-médias genético (SOMGK) \cite{nath:Kuo2005}. O AK é um método de agrupamento robusto, porém, apresenta ainda alguns problemas os quais dizem res\-pei\-to ao número de agrupamentos, que é requerido \emph{a priori} da mesma forma que o algoritmo K-médias convencional.

O sistema híbrido SOMK é composto por 2 Níveis de Abstração. O primeiro nível é a rede SOM e o segundo o algoritmo K-médias. A Figura ~\ref{fig:OsDoisNiveisAbstracao} apresenta o funcionamento deste sistema híbrido de forma mais bem definida, lembrando que o segundo nível é o algoritmo de agrupamento K-médias. Assim, é indispensável substituir este método de agrupamento com a finalidade de suprir deficiências como: saber \emph{a priori} o número de agrupamentos do algoritmo K-médias e informar os centróides iniciais com SOMK. Onde, o \emph{k} inicial é igual ao número de classes da base de dados a ser usada.

A seleção de um bom modelo de previsão de uma série temporal é uma tarefa que envolve experiência e conhecimento.  Sistemas híbridos foram utilizados com sucesso para combinar previsões de séries temporais. O sistema híbrido usa recursos das séries em questão para definir pesos adequados para os métodos de previsão individuais que estão sendo combinados. As previsões combinadas são a média ponderada das previsões fornecidas pelos métodos individuais \cite{Ricardo-australia-2004}. O método hibrido conseguiu obter previsões significativamente precisas. Uma outra solução utilizando sistema híbrido foi aplicado na tarefa de selecionar os melhores modelos de previsão entre modelos amplamente difundidos na literatura \cite{Ricardo-PRL-2004}. 

\subsubsection{Algoritmo de Agrupamento baseado nas Colônias de Formigas (ASCA)}

\begin{figure}[h!]
\begin{center}
\includegraphics[width=7cm,height=12cm]{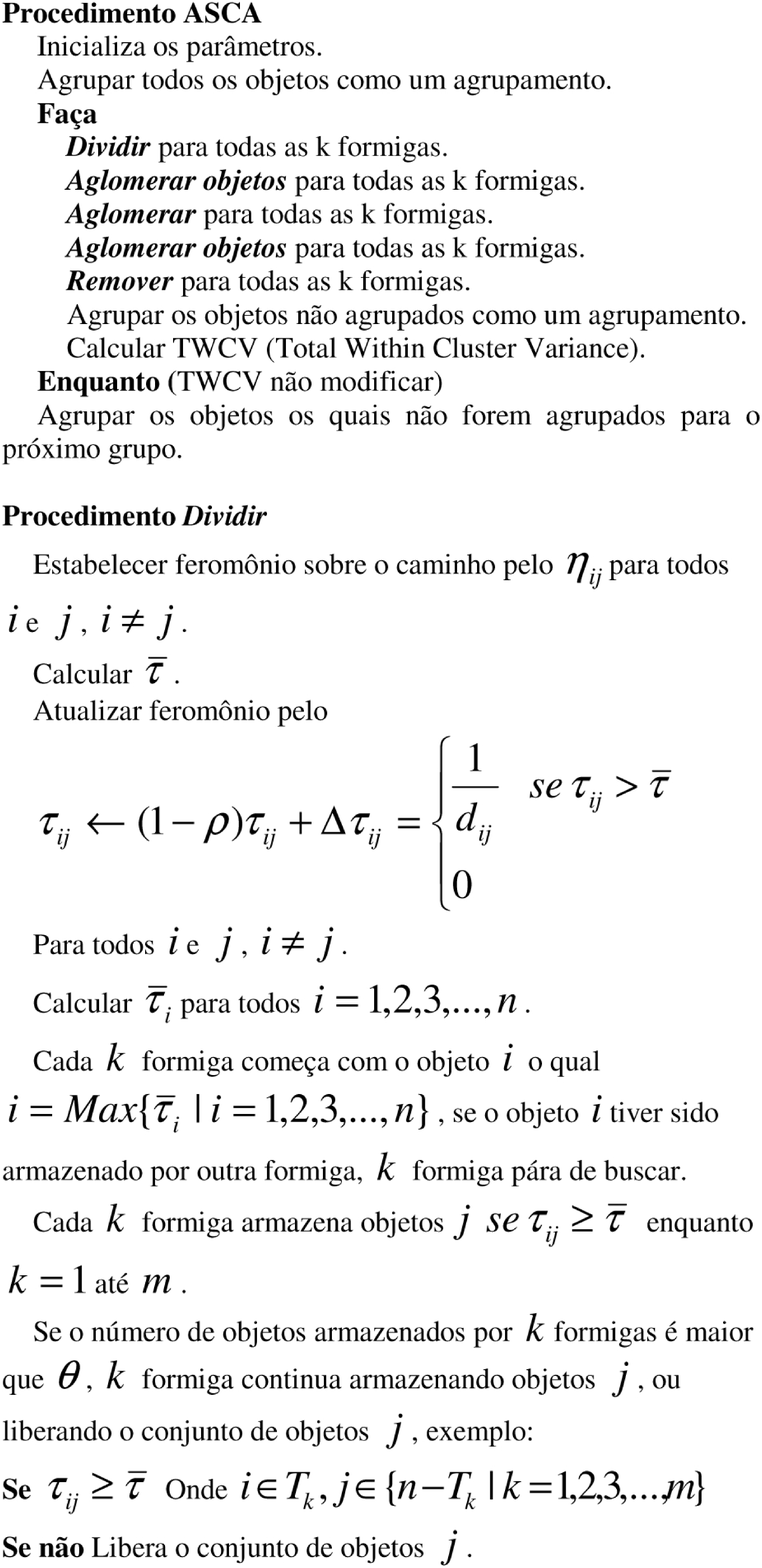}
\end{center}
\vspace{-15pt}
\caption{O procedimento ASCA.}
\label{fig:ProcedimentoASCA01}
\end{figure}

\begin{figure}[h!]
\begin{center}
\includegraphics[width=7cm,height=12cm]{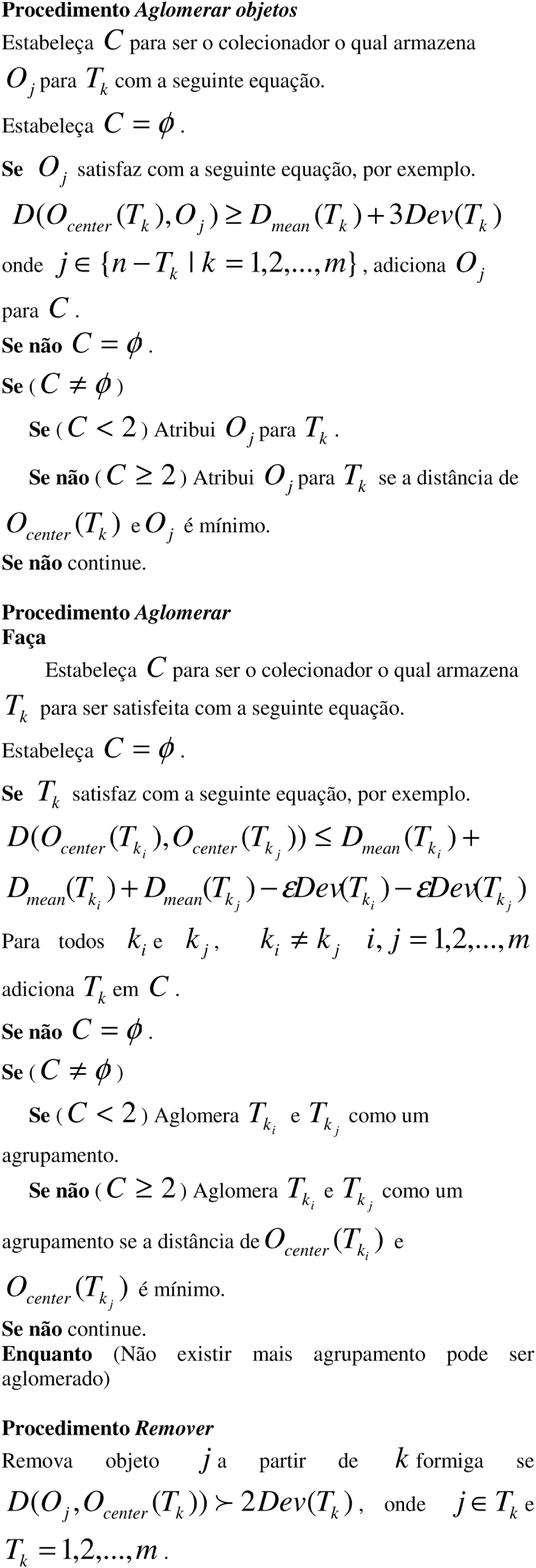}
\end{center}
\vspace{-15pt}
\caption{Continuação do procedimento ASCA.}
\label{fig:ProcedimentoASCA02}
\end{figure}

Kuo et al. \cite{nath:Kuo2005} propuseram um Algoritmo de Agrupamento baseado em Colônias de Formigas (ASCA). ASCA possui quatro subprocedimentos, que são: \emph{dividir}, \emph{a\-glo\-me\-rar objetos}, \emph{aglomerar} e \emph{remover}. A seguir, o procedimento ASCA será descrito em detalhes.  Primeiro o algoritmo inicializa os parâmetros e agrupa todos os objetos ou dados em um único agrupamento, então, no segundo passo do algoritimo, o sub-procedimento \emph{dividir} irá dividir este único agrupamento em sub-agrupamentos. Terceiro, depois de \emph{dividir}, a próxima etapa deste algoritmo ASCA  é a execução do sub-procedimento \emph{aglomerar objetos}; com o propósito de agrupar os objetos dentro de sub-agrupamentos adequados. Em quarto lugar, o sub-procedimento \emph{aglomerar} integra dois sub-procedimentos iguais em um único agrupamento. Logo em seguida, executa o sub-procedimento \emph{aglomerar objetos} novamente. Depois de agrupar os objetos similares dentro de sub-procedimentos adequados, o sub-procedimento \emph{remover} excluirá objetos que não são similares. Desse modo, calcula o total de variância, TWCV, dentro de cada agrupamento. Se TWCV não for modificado, agrupa os objetos que ainda não foram agrupados para o próximo agrupamento e pára o algoritmo ASCA. Do contrário, repete os sub-procedimentos na seguinte seqüência: \emph{dividir}, \emph{aglomerar objetos}, \emph{a\-glo\-me\-rar}, \emph{aglomerar objetos} e \emph{remover}, até quando TWCV não for modificado. O algoritmo ASCA \cite{nath:Kuo2003} podem ser encontrados nas Figuras~\ref{fig:ProcedimentoASCA01} e ~\ref{fig:ProcedimentoASCA02}.

\vspace{-10pt}

\subsubsection{\emph{Ant} K-médias}

A idéia do algoritmo \emph{Ant} K-médias (AK) é realizar os agrupamentos finais dos modelos SOM, ASCA e SOINN. Uma vez que SOM, ASCA e SOINN são aplicados no primeiro nível de abstração encontrando os protótipos ou centróides. Existe também, o procedimento pertubação dentro de AK que é útil para saltar a solução do minímo local. O algoritmo AK produziu resultados significativos, no que diz respeito ao problema de agrupamento. Então, o objetivo inicial foi usar o método de agrupamento que é formado em dois estágios (SOMAK), a fim de obter a taxa de erro baseada na entropia e o tempo computacional menor quando comparado com SOMK. Neste método, é necessário fornecer o número de agrupamentos tal como o algoritmo K-médias convencional. Os termos e notações abaixo serão usados durante o algoritmo. Seja

\vspace{-25pt}
\begin{center}
    \begin{equation}
         E = {O_{1}, O_{2}, ..., O_{n}}
    \end{equation}
\end{center}

o conjunto de \textit{n} dados (ou objetos), onde \textit{O} são objetos (dados, padrões ou amostras) colecionados a partir da base de dados, em que cada objeto tem \textit{A} atributos, onde \textit{A} $ > 0 $, como mostrado na Tabela ~\ref{tab:FormatoConjuntoDados}. Alguns parâmetros importantes, como: $\alpha$: a importância relativa da trilha ($\alpha >= 0$), $\beta$: a importância relativa da visibilidade ($\beta >= 0$), $\rho$: o parâmetro de decaimento do feromônio ($0 < \rho < 1$), \textit{Q}: uma constante ($\emph{Q} > 0$), \textit{n}: número de objetos, \textit{m}: número de formigas, \textit{nc}: número de agrupamentos e \textit{T}: conjunto que inclui objetos usados, onde o número máximo armazenado pelo vetor \textit{T} será de \textit{n}, ou seja,

\vspace{-20pt}
\begin{center}
    \begin{equation}
         T = {O_{a}, O_{b}, ..., O_{t}}
    \end{equation}
\end{center}

onde \textit{a, b, ..., t} são pontos que as formigas têm visitado. \textit{$T_{k}$}: conjunto \textit{T} construído por \textit{k} formigas e \textit{$O_{center}(T) $}: centro de todos os objetos em \textit{T}, onde \textit{nT} representa o número de objetos em \textit{T}, ou seja,

\vspace{-20pt}
\begin{center}
    \begin{equation}
        O_{center}(T) = \frac{1}{nT} \sum_{i=1}^{n} O_{i}
    \end{equation}
\end{center}

\textit{TWCV}: variância total dentro do agrupamento, ou seja,

\vspace{-20pt}
\begin{center}
    \begin{equation}
        \sum_{k=1}^{nc} \sum_{i=1}^{n} (O_{i} - O_{center(T_{k})})^{2}
    \end{equation}
\end{center}
\vspace{-5pt}

\vspace{-10pt}
\setlength{\tabcolsep}{0.95pt}
\begin{table}[h!]
\begin{small}
\begin{center}
  \caption{A representação do conjunto de dados \cite{nath:Kuo2005}}
  \vspace{-8pt}
  \label{tab:FormatoConjuntoDados}
  \begin{tabular} {l*{4}{c}r} \\ \hline
    Objects & $ A_{1} $ & $ A_{1} $ & \ldots & $ A_{1} $ \\ \hline
    $ O_{1} $ & 1.22 & 32.5 & \ldots & 56.4 \\
    \vdots & \vdots & \vdots & \ldots & \vdots  \\
    $ O_{n} $ & 55.6 & 5.6 & \ldots & 8.4 \\ \hline
  \end{tabular}
\end{center}
\end{small}
\vspace{-25pt}
\end{table}

\vspace{5pt}

A primeira etapa do algoritmo é a inicialização dos parâmetros, isto é, a inclusão do número de agrupamentos e seus respectivos centróides. Depois, é necessário estabelece igualdade de feromônio para cada caminho e na terceia etapa, cada \textit{k} formiga escolhe o centróide mais perto para se mover com a probabilidade \textit{P}.

\vspace{-25pt}
\begin{center}
    \begin{equation}
        P_{\imath\jmath}^{k} = \dfrac{\tau_{\imath\jmath}^{\alpha} \eta_{\imath\jmath}^{\beta}} {\sum_{c}^{nc} \tau_{\imath c}^{\alpha} \eta_{\imath c}^{\beta}}
    \end{equation}
\end{center}

onde $ \imath $ representa o ponto inicial e $ \jmath $ o final (centróide) para onde a \textit{k} formiga escolhe se mover, \textit{c} é o centróide e \textit{nc} é o número de centróides. No entanto, se o valor de \textit{$ P_{\imath\jmath} $} é maior do que os outros, a \textit{k} formiga se moverá do ponto $ \imath $ para $ \jmath $, e o objeto $ \imath $ pertence ao centróide $ \jmath $. Finalmente, o feromônio é atualizado pela Equação (8).

\begin{figure}[h!]
\begin{center}
\includegraphics[height=7.2cm]{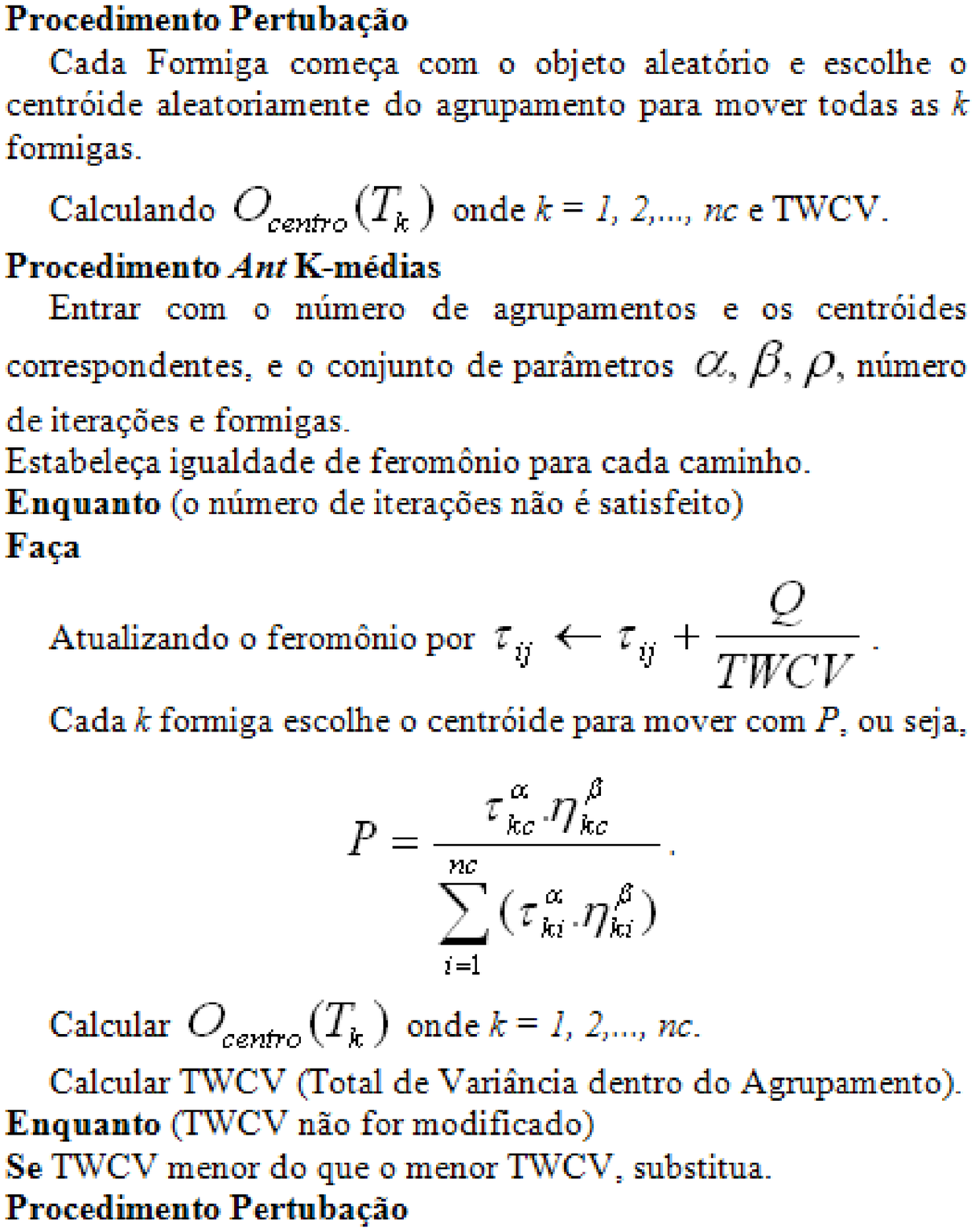}
\end{center}
\vspace{-20pt}
\caption{O procedimento \emph{Ant} K-médias \cite{nath:Kuo2005}}
\label{fig:ProcedimentoAntKmeans}
\end{figure}

\vspace{-20pt}
\begin{center}
    \begin{equation}
        \tau_{\imath\jmath} \leftarrow \tau_{\imath\jmath} + \dfrac{Q}{TWCV}
    \end{equation}
\end{center}

onde \textit{Q} representa uma constante, \textit{TWCV} é total de variância dentro do agrupamento. Então, é necessário calcular o centro de todos os objetos em \textit{T}, onde \textit{k = 1, 2, 3, ..., nc}. Depois, calcular \textit{TWCV}. Por último, se \textit{TWCV} for alterado, volta-se para a terceira etapa; caso contrário, se \textit{TWCV} é menor do que o menor \textit{TWCV}, substitua. Em seguida, execute o procedimento pertubação afim de saltar a solução do minímo local. Se o número de iterações não é atingido, faz-se necessário voltar para terceira etapa; caso contrário, pare o algoritmo. Figura ~\ref{fig:ProcedimentoAntKmeans} mostra o procedimento \emph{Ant} K-médias.

\subsection{Parâmetros}
Os parâmetros considerados neste trabalho afetam as técnicas de agrupamento, citadas anteriormente para solucionar o problema de agrupamento. De acordo com \cite{nath:Dorigo2000}, existem diversas combinações para determinar os parâmetros aplicados ao algoritmo de colônia de formigas. Normalmente, os parâmetros são $\alpha$ = {0, 0.5, 1, 2, 5}; $\beta$ = {0, 1, 2, 5}; $\gamma$ = {0.3, 0.5, 0.7, 0.99, 0.999} e Q = {1, 100, 10000}. Existem 300 combinações de parâmetros, os resultados mostraram em \cite{nath:Berkhin2009}, que $\alpha$ = 0.5 , $\beta$ = 1, $\rho$ = 0.9 e Q = 1 apresentam a menor variância para dados estatísticos. A Tabela ~\ref{tab:ParametrosPrincipaisTecnicasAgrupamentos} mostra os parâmetros das técnicas de agrupamento.

\setlength{\tabcolsep}{0.95pt}
\begin{table}[h!]
\begin{small}
\begin{center}
  \caption{Parâmetros das Técnicas de Agrupamento}
  \vspace{-10pt}
  \label{tab:ParametrosPrincipaisTecnicasAgrupamentos}
  \begin{tabular} {l*{2}{c}r} \\ \hline
    Técnicas & Parâmetros \\ \hline
        &  número de atributos = quantidade dos padrões de entrada, \\
        &  taxa de aprendizagem inicial = 0.5 e final = 0.99, \\
        &  tamanho das linhas do mapa = 19 e colunas = 17, \\
    SOM &  raio inicial = 10 e final = 2, função de vizinhança = gaussiana, \\
        &  formato da vizinhança = hexa, tipo de treinamento = épocas, \\
        &  tamanho do treinamento para fase \emph{rough} = 3 e fase \emph{fine-tuning} = 10 \\ \hline
    SOINN & limiares de similaridade, TAA e fatores de decaimento de E, M e R \\ \hline
    K-médias &  k = número de agrupamentos iniciais e inicialização (ini.) dos k centros \\ \hline
    ASCA &  $ \epsilon = rand(0,1) $ e $ \theta = 1000 $ \\ \hline
    SOMK & k = nº de protótipos SOM e ini. dos centros = número de centróides SOM \\ \hline
  \end{tabular}
\end{center}
\end{small}
\end{table}

\section{Métodos Propostos}
\subsection{Arquitetura Geral}
\begin{figure}[h!]
\begin{center}
\includegraphics[width=6cm,height=4cm]{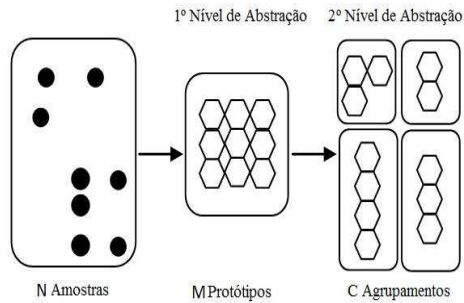}
\end{center}
\vspace{-20pt}
\caption{Primeiro nível é obtido através da criação de um conjunto de vetores de protótipos usando SOM, ASCA e SOINN. Os métodos propostos SOMAK, ASCAK e SOINAK criam o segundo nível rea\-li\-zan\-do o agrupamento dos M protótipos.}
\label{fig:OsDoisNiveisAbstracao}
\end{figure}

Neste trabalho, os métodos propostos baseado em dois estágios são relevantes para melhorar as principais deficiências de um método baseado em um estágio (e.g. K-médias), que tem se mostrado sensível aos protótipos iniciais e à determinação de um número apropriado de \emph{k} agrupamentos. O funcionamento de um método baseado em dois estágios tem como propósito suprir as dificuldades que uma técnica de agrupamento apresenta individualmente. Assim, pesquisadores em \cite{nath:Vesanto2000} têm combinado duas ou mais técnicas distintas para solucionar um dado problema de agrupamento. Primeiro, um conjunto grande de protótipos é formado usando as redes neurais não-supervisionada (SOM e SOINN) e o algoritmo ASCA. Os protótipos podem ser interpretados como \textit{proto-agrupamentos} que são, na próxima etapa, combinados para formar os agrupamentos finais. Cada vetor de dados do conjunto de dados original pertence ao mesmo agrupamento com seu protótipo mais próximo. A Figura ~\ref{fig:OsDoisNiveisAbstracao} retrata a arquitetura do método de agrupamento baseado em dois estágios \cite{nath:Vesanto2000}, mostrando os dois níveis de abstração. O primeiro níıvel de abstração pode ser (a rede neural auto-organizável, o algoritmo ASCA e a rede neural incremental auto-organizável) e o segundo níıvel de abstração é e o algoritmo Ant K-médias, o qual tem a responsabilidade de formar os agrupamentos finais ou definitivos.

Considere \textit{N} (amostras de dados) utilizando o algoritmo Ant K-médias, AK, (Figura ~\ref{fig:ProcedimentoAntKmeans}). Isto implica fazer, tentativas de agrupamentos com diversos valores para o número de protótipos que foram obtidos pela rede SOM, o algoritmo ASCA e a rede incremental SOINN. A complexidade computacional é proporcional a $\sum_{k=2}^{C_{max}} \emph{Nk}$, onde $ C_{max} $ é o número máximo pré-estabelecido de agrupamentos e \textit{k} representa o número de agrupamentos iniciais. Quando um conjunto de protótipos é usado em uma etapa intermediária (Figura ~\ref{fig:OsDoisNiveisAbstracao} - 1º Nível de Abstração), a com\-ple\-xi\-da\-de é proporcional a $ NM + \sum_{k=2}^{M} \emph{Mk}$, onde \textit{M} é o número de protótipos obtidos. Com $ C_{max} = \sqrt{N} $ e $ M = 5\sqrt{N} $, a redução do custo computacional é baseada $ \dfrac{\sqrt{N}}{15}$, ou cerca de seis-\textit{folds} para $ N = 1000 $ amostras de dados \cite{nath:Vesanto2000}. Neste trabalho, usamos dez-\textit{folds} e $ N = 1000 $ para a realização dos experimentos. Evidentemente que esta é uma estimativa, e considerações práticas e experimentais são ignoradas.

\subsection{SOMAK}
Pelo fato do algoritmo AK apresentar resultados significativos quando comparado com os métodos de agrupamento SOMK e SOMGK, Souza et al. propuseram o algoritmo SOMAK \cite{nath:Souza2009a} \cite{nath:Souza2009b} com o propósito de realizar uma análise de agrupamento dos dados e, assim, obter resultados mais significativos do que os já apresentados na literatura. A principal vantagem do sistema híbrido SOMAK está em sua fase inicial, ou seja, o pré-processamento dos dados apresentados para o algoritmo AK. Tal pré-processamento é feito pela rede SOM, enfatizando a utilização de AK no segundo nível, SOMAK tem mostrado resultados significativos. A arquitetura deste método de agrupamento pode ser vista na Figura ~\ref{fig:OsDoisNiveisAbstracao}. SOMAK utiliza SOM como método de primeiro estágio sobre os dados de entrada, ao invés de agrupar os dados diretamente.

No estudo atual, o número de agrupamentos e o centróide de cada agrupamento são gerados a partir da rede SOM, a fim de validar a solução de análise de agrupamento. SOMAK usa SOM com o propósito de determinar o número de agrupamentos e os pontos iniciais, depois, aplica o procedimento AK para encontrar a solução final.

\subsection{ASCAK}
Com a crescente busca de melhoria no sistema híbrido para o reconhecimento de padrões e a necessidade de substituir o método de agrupamento SOMAK com a perspectiva de se obter resultados mais promissores, um novo método de agrupamento, baseado em dois estágios, foi aplicado combinando os algoritmos ASCA e o AK (ASCAK). O algoritmo ASCA \cite{nath:Kuo2005} possui basicamente quatro sub-procedimentos, que são: \emph{dividir}, \emph{aglomerar objetos}, \emph{aglomerar} e \emph{remover}, descrito nas Figuras ~\ref{fig:ProcedimentoASCA01} e ~\ref{fig:ProcedimentoASCA02}. Assim, deve-se ressaltar que não encontrará um agrupamento ótimo para os dados, mas obterá uma visão sobre a estrutura de agrupamento dos dados, usando o sistema híbrido ASCAK, com a finalidade de reduzir o tempo computacional e o de\-sem\-pe\-nho quando comparado com outras técnicas de agrupamento (SOMK e SOMAK).

Realizamos uma pesquisa criteriosa na literatura e observamos que o algoritmo ASCA tem mostrado resultados melhores quando comparado com SOM \cite{nath:Kuo2003}. Assim, ASCAK foi proposto neste trabalho e a única diferença entre os métodos de agrupamentos SOMAK e o ASCAK é à fase de pré-processamento dos dados, em que o algoritmo ASCA substituiu SOM.

\subsection{SOINAK}
A rede neural SOINN tem duas maneiras possíveis de funcionamento. Na primeira, a rede neural incremental SOINN é inicialmente utilizada para determinar o número de grupos e, na segunda, os centros dos grupos iniciais são informados para AK. O centro inicial de um grupo pode ser obtido a partir do vetor peso correspondente aos centros dos grupos sobre a topologia de SOINN. O método utilizado no segundo estágio é o AK, e a principal vantagem de um método de dois estágios baseado em SOINN é diferenciada quando comparada à combinação dos modelos apresentados (SOMK, SOMAK e ASCAK), isto é, o fato que não é necessário saber, \emph{a priori}, o número de agrupamentos de K-medias (ou AK) e outro benefício importante é que não necessita informar os centróides iniciais.

Visando sempre um sistema híbrido de reconhecimento de padrões aplicado ao pro\-ble\-ma de agrupamento o mais automático possível e a redução da quantidade de parâmetros para que o usuário realize a configuração mais adequada, visto que, é necessário que o mesmo configure uma série de parâmetros para serem informados a rede SOM, ficando sem uma confiança adequada de melhor configuração da rede. Em virtude de todas essas deficiências apresentadas, Furao e Hasegawa \cite{nath:Furao2006} propuseram SOINN, com o intuito de re\-a\-li\-zar uma aprendizagem não-supervisionada dos dados e suprir estas deficiências.

O SOINN se propõe a ser uma rede de representação topológica, que gera grupos de dados correlacionados e que podem ser capazes de realizar aprendizado de forma incremental. SOINN usa uma quantidade de parâmetros menor, quando com\-pa\-ra\-da à SOM, onde quanto maior for a quantidade de parâmetros para o usuário configurar, menor será a chance deste sistema híbrido a\-pre\-sen\-tar resultados significativos. Portanto, SOINAK busca obter melhores resultados. Tal qual o ASCAK, a diferença que existe entre o SOMAK e o SOINAK é a fase de pré-processamento dos dados, visto que a rede SOINN susbstituiu SOM.

\subsection{Considerações}
O objetivo deste trabalho foi propor os métodos de agrupamentos SOMAK, ASCAK e SOINAK, criado como dois níveis de abstrações, utilizado no agrupamento dos dados. O primeiro nível de abstração usado no agrupamento de dados tem algumas van\-ta\-gens. O conjunto de dados original é representado através de um conjunto menor de vetores protótipos, permitindo uma utilização eficiente dos algoritmos de agrupamento para dividir os protótipos em grupos. Depois, aplicar redes neurais auto-organizáveis e o ASCA possibilitando apresentação visual e a interpretação dos agrupamentos. A Tabela ~\ref{tab:ParametrosPrincipaisMetodosPropostos} mostra os parâmetros principais dos métodos propostos deste trabalho.

\setlength{\tabcolsep}{0.95pt}
\begin{table}[h!]
\begin{small}
\begin{center}
  \caption{Parâmetros principais dos Métodos Propostos \cite{nath:Berkhin2009}}
  \vspace{-10pt}
  \label{tab:ParametrosPrincipaisMetodosPropostos}
  \begin{tabular} {l*{2}{c}r} \\ \hline
    Técnicas & Parâmetros \\ \hline
    SOMAK & $\alpha = 0.5$, $\beta = 1$, $\gamma = 0.9$, Q = 1, n = 500, m = 2 e nc = nº de protótipos SOM \\ \hline
    ASCAK & $\alpha = 0.5$, $\beta = 1$, $\gamma = 0.9$, Q = 1, n = 500, m = 2 e nc = nº de protótipos ASCA \\ \hline
    SOINAK& $\alpha = 0.5$, $\beta = 1$, $\gamma = 0.9$, Q = 1, n = 500, m = 2 e nc = nº de protótipos SOINN \\ \hline
  \end{tabular}
\end{center}
\end{small}
\end{table}

\section{Materiais e Métodos}
\subsection{Conjunto de Dados}
Esta seção descreve os conjuntos de dados usados neste trabalho, que são: dados sintéticos e dados reais. Estes conjuntos são úteis para comprovar que os métodos propostos são realmente robustos, podendo ser aplicáveis a diferentes problemas de agrupamento, visto que, abordamos os diferentes dados.

A escolha dos dados sintéticos ocorreu utilizando a ferramenta \emph{PRTools4 A Matlab Toolbox for Pattern Recognition} \cite{nath:Duin2007}, que fornece controle e de\-ta\-lhes destas bases, visando atingir problemas simples e complexos. Quanto as bases reais extraídas do repositório UCI \cite{nath:Aha1987} usamos as bases \emph{Balance Scale} e \emph{Contraceptive Method Choice} para teste experimental, onde queríamos observar como os métodos propostos se comportaríam com estas bases de três classes ou grupos. Já com relação às demais bases reais usamos por serem encontradas na literatura para feitos de agrupamento. Os dados possuem baixa dimensão, várias instâncias e poucos grupos. 
%Por fim, com 10 bases apenas não é suficiente para se obter uma conclusão, é preciso ter mais dados, por exemplo, para contemplar a relação de dimensionalidade (alta e baixa dimensão) que não foi vista neste trabalho.

\subsubsection{Dados Sintéticos}
Dados gerados segundo regras que permitam garantir determinadas propriedades nos dados, validando sua resposta em situações de teste controlado, para que os métodos de agrupamentos implementados possam ser aferidos. Os dados são gerados pelo método Monte Carlo \cite{nath:Milligan1980} e \cite{nath:Milligan1985} visando validar a viabilidade dos métodos e termos uma maior confiabilidade nos dados. Assim, a taxa de erro e a variância dentro dos agrupamentos podem ser obtidas de forma a validar as soluções dos métodos de agrupamentos.

Da segunda base de dados em diante, \textit{A} representa os dados que são ge\-ra\-dos pela ferramenta \emph{PRTools4} e \textit{N} o número de amostras que foram obtidas pelo vetor, com o número de amostras por classe. $ N = [500, 500] $ representa 1000 padrões de dados. A seguir serão descritas as cinco bases de dados sintéticas. A base de dados \emph{Lines} possui 1000 padrões, que foram obtidos através do Método Monte Carlo, que agrupou em sua solução final 10 segmentos, ou seja, 10 grupos iniciais para o conjunto de dados. Na base de dados \emph{Banana}, \textit{A} representa duas dimensões e duas classes de \textit{N} objetos, apresentando uma distribuição em forma de banana. Assim, seus respectivos dados (padrões) são distribuídos uniformemente ao longo de uma banana e está sobreposta com uma distribuição normal de desvio padrão \textit{S}, em todas as direções \textit{S = 1}. Na base de dados \emph{Highleyman} a geração do conjunto de dados \textit{A} representa duas dimensões e duas classes de \textit{N} objetos definidas como: (a) Geração de 500 padrões de dados para cada uma das duas distribuições com média 1 e 0 e variâncias 0 e 0.25 de uma dimensão (Gaussiana) e (b) Geração de 500 padrões de dados para cada uma das duas distribuições com média 0.01 e 0 e variâncias 0 e 4 de uma dimensão (Gaussiana).

O \emph{Simple Problem} (problema simples de classificação) apresenta duas dimensões e duas classes. As classes têm distribuições gau\-ssi\-a\-nas com uma matriz identidade como matriz de covariância. As médias, se encontram numa distância \textit{D} (distância entre as médias das classes na primeira dimensão). A base \emph{Spherical} contém duas classes esféricas com variâncias diferentes. As duas classes têm distribuições gaussianas em forma de uma esfera, Classe 1 representa uma matriz identidade como matriz de covariância e média \textit{U}. Se a média \textit{U} é um escalar, então [U, 0, 0, ..], assim, sendo útil como média de classe e Classe 2 representa da mesma forma que a Classe 1, exceto para uma variância de 4 para o primeiro de duas características com média 0.

\subsubsection{Dados Reais}

Os dados reais usados neste artigo foram obtidos do repositório UCI \cite{nath:Aha1987} e são: \emph{Balance Scale}, \emph{Contraceptive Method Choice}, \emph{Dermatology}, \emph{Diabetes} e \emph{Glass}.

\begin{enumerate}
  \item \emph{Balance Scale}: Dados gerados como resultado experimental de modelos psicológicos, onde o exemplo é classificado como tendo o equilíbrio da escala para a ponta direita, ponta esquerda ou ser equilibrado. A base de dados é formada por 625 registros com 4 atributos e que correspondem às informações da escala de equilíbrio (balança). A análise dos dados foram realizadas a partir de um algoritmo que gerou resultados entre 1 e 3, sendo assim: (a) valor 1 representa que a escala está balanceada com 49 registros (7.84\%), (b) valor 2 representa que a escala está à esquerda com 288 registros (46.08\%) e (c) valor 3 representa que a escala está à direita com 288 registros (46.08\%).
  \item \emph{Contraceptive Method Choice} (CMC): Método contraceptivo (sem uso, métodos de longo prazo e métodos a curto prazo, são os atributos de classe) de uma mulher com base nas suas ca\-ra\-cte\-rís\-ti\-cas econômicas e sócio-demográficas. Estes dados apresentam um subconjunto da pesquisa nacional de métodos contraceptivos da indonésia em 1987, as amostras são mulheres casadas que não estão grávidas ou não estavam no momento da entrevista. Os dados são formados por 1473 registros com 9 atributos de entrada e um atributo de classe que corresponde as informações de um método contraceptivo.
  \item \emph{Dermatology}: A base de dados é formada por 366 registros com 33 atributos e correspondem à informações em um problema real de dermatologia. A análise dos dados foram realizadas a partir de um algoritmo que gerou re\-sul\-ta\-dos entre 1 e 6, sendo assim: (a) valor 1 indica que o paciente possui psoríase com 112 registros (30.60\%), (b) valor 2 indica que possui dermatite seborréica com 61 registros (16.66\%), (c) valor 3 indica que possui o plano líquen com 72 registros (19.67\%), (d) valor 4 indica que possui pitiríase rósea com 49 registros (13.38\%), (e) valor 5 indica que possui dermatite crônica com 52 registros (14.20\%) e (f) valor 6 indica que possui pitiríase rubra pilar com 20 registros (5.49\%).
  \item \emph{Diabetes}: Base conhecida na UCI como \emph{Pima Indians Diabetes Data Set}, extraída de dados pessoais e exames médicos de pacientes. Usada para a classificação de uma índia Pima ser diabética. Todos os pacientes são mulheres com a idade mínima de 21 anos, USA. Os dados são formados por 768 registros com 8 atributos contínuos e que correspondem às informações médicas. A análise dos dados foram realizadas a partir de um algoritmo, que gerou resultados contínuos, que foram transformados em valores binários entre 0 e 1 com valor de corte = 0.448, sendo assim: (a) valor 0 paciente não possui diabetes com 500 registros (65.1\%) e (b) valor 1 possui diabetes com 268 registros (34.9\%).
  \item \emph{Glass}: Base ex\-tra\-í\-da a partir dos serviços de ciência da \emph{Forensic}, \emph{USA}. Nestes dados foi realizado um teste comparativo de um sistema baseado em regras BEAGLE, através do algoritmo de vizinho mais próximo e análise discriminante, com o objetivo de determinar se o vidro pertence a um tipo "\emph{float}" ou não, os resultados obtidos foram: janelas processadas com o tipo "\emph{float}" (87) e janelas que não foram processadas com o tipo "\emph{float}" (76). Esta base é formada por 214 registros com 11 atributos contínuos (incluindo o atributo de classe) e que correspondem às informações de vidro.
\end{enumerate}

\subsection{Pré-processamento}
A técnica de pré-processamento de dados é um processo de transformação dos dados de entrada em dados mais específicos, para serem usados para a extração de características e eliminação de ruídos. Para que o treinamento das redes neurais SOM e SOINN sejam eficientes e tenham uma maior ra\-pi\-dez, é aconselhável fazer uma transformação dos dados para um intervalo bem definido. Os dados com atributos diferentes podem fazer uma confusão no aprendizado da rede, resultando maiores considerações para valores com grandes magnitudes. A técnica de normalização evita este problema, transpondo os dados em intervalos bem definidos. Existem diferentes normalizações, tais como: logarítmicas, tangente hi\-per\-bó\-li\-ca, entre outras. Vários são os tipos de normalizações, depende do objetivo que se pretende atingir com as RNAs. Os intervalos comumente vistos são entre 0 a 1 ou -1 a 1. Neste trabalho, os experimentos foram realizados com os padrões (sintéticos e reais) normalizados, usando a técnica de normalização para que os valores ficassem dispostos entre -1 e 1; onde $ x_{norm} $ é o valor normalizado correspondente ao valor original $ x $. $ x_{min} $ e $ x_{max} $ são os valores de mínimo e máximo entre os padrões de entrada das RNAs.

\vspace{-20pt}
\begin{center}
    \begin{equation}
        x_{norm} = \dfrac{(2x - x_{max} - x_{min})}{(x_{max} - x_{min})}
    \end{equation}
\end{center}

\subsection{Divisão dos padrões em Treino e Teste}
Para solucionar o problema de agrupamento através da utilização das redes neurais SOM e SOINN com o treinamento não-supervisionado, não se faz necessário a utilização dos dados que correspondam a pares de entrada-saída já conhecidos. A divisão dos conjuntos em treinamento e teste não tem a ver com o tamanho do conjunto e sim com a necessidade de um conjunto independente testar o desempenho da rede. O treinamento das RNAs são realizados com os dados de treino. Assim, é importante medir o desempenho da rede com um conjunto de dados independente, conhecido como conjunto de teste. A avaliação feita a partir do conjunto de teste mede o grau de generalização da rede neural para o problema abordado.

A técnica de validação cruzada (do inglês, \emph{cross-validation}) foi desenvolvida na ferramenta MATLAB \cite{nath:Moler1970}, usando o conceito básico da validação cruzada, que é dividir o conjunto de dados original em dados de treinamento e teste. Neste trabalho foram realizados 10 partições de dados (10-\emph{folds}), cada uma destas partições consiste da mesma base de dados, porém com suas instâncias dispostas em ordens di\-fe\-ren\-tes (aleatórias). Por exemplo, um classificador é treinado usando um subconjunto formado pela junção de 9 destes grupos e testado usando o grupo remanescente. Isto é feito 10 vezes, e cada vez um grupo diferente é usado como conjunto de teste, e computa-se o erro do conjunto de teste, $E_{i}$. Finalmente, o erro da validação cruzada é computado como a média sobre os 10 erros $E_{i}$, $1\leq i\leq10$. O uso da validação cruzada ajuda a eliminar o viés dos resultados. 

Com relação aos 10-\emph{folds} \emph{cross-validation}, um padrão do conjunto de teste é atribuído a cada grupo. Primeiro, é realizado o cálculo do centróide através da distância euclidiana, onde cada padrão é avaliado com as classses ou grupos informados das bases de dados. Segundo, é atribuído o padrão ao centróide mais próximo dos grupos (calcula-se a distância euclidiana entre o padrão e os centróides dos grupos), ou seja, o padrão pertencerá ao grupo que apresentou a menor distância entre o padrão e todos os centróides dos grupos, assim os elementos do conjunto de teste são atribuídos aos grupos.

\subsection{Aspectos observados}
Os métodos usados para solucionar o problema de agrupamento foram implementados na linguagem MATLAB. SOM usou o algoritmo de treinamento seqüencial para os conjuntos de dados \footnote{No treinamento da rede neural SOM, o pacote do Matlab disponível livremente SOM Toolbox foi utilizada. Para mais informações, ver URL http://www.cis.hut.fi/projects/somtoolbox/ \cite{nath:Alhoniemi2005}}.

A definição dos \emph{k} centróides iniciais para os métodos de agrupamentos neste trabalho é igual ao número de classes das bases de dados iniciais com o propósito de solucionar o problema de agrupamento de dados.

Para medir a acurácia dos métodos de agrupamentos neste trabalho, foi usada a medida de entropia. A motivação para o uso da entropia como medida de desempenho surge da capacidade que esse conceito estatístico possui de medir o grau de aleatoriedade (ou de incerteza) dos dados, além disso, esta medida tem sido aplicada para avaliar os métodos de agrupamentos em dois estágios \cite{nath:Vesanto2000}, também foram usados para medir o grau de desordem de um determinado método de agrupamento e finalmente, avaliar a qualidade de um método. 

Antes de encontrar a taxa de erro baseada na entropia, foi feita uma verificação dos \emph{labels} dos protótipos obtidos dos métodos de agrupamentos com os \emph{labels} dos padrões da base de dados. Se for classificado corretamente, ou seja, se os \emph{labels} dos protótipos é igual aos \emph{labels} dos padrões, então incrementa o valor de $m_{i}$, caso contrário, incrementa o valor de $m_{ij}$. %Estes valores ($m_{i}$ e $m_{ij}$) serão descritos em detalhes posteriormente.

Quando temos informações externas dos dados, tipicamente em forma de rótulos de classe derivados dos dados; o procedimento usual é medir o grau de correspondência entre os rótulos de \emph{cluster} e os de classe. Visando os rótulos de classe, faz-se necessário descobrir o foco em realizar uma Análise de \emph{Cluster} (AC). As motivações são de realizar comparações entre técnicas de agrupamento ou avaliar o grau em que um processo de classificação manual pode ser automaticamente produzido pela AC.

Dois diferentes tipos de abordagens são citados em \cite{nath:Tan2006}. O primeiro conjunto de medidas de desempenho usados para avaliar as técnicas com o objetivo de solucionar o problema de agrupamento são: entropia, pureza e a \emph{F-measure}. Estas medidas avaliam a extensão a qual um \emph{cluster} contém objetos de uma única classe. O segundo grupo de medidas é relacionado com a medida de similaridade para dados binários, como a medida de \emph{Jaccard}. Este grupo de medidas de desempenho abordam a extensão a qual dois objetos que estão na mesma classe são do mesmo \emph{cluster} e virce versa. Para conveniência, estes dois tipos de medidas são referenciados como orientado a classificação e a similaridade respectivamente.

A medida de erro usada na análise dos resultados é a entropia, onde para cada agrupamento, a distribuição de classe de dados é primeiramente calculada, ou seja, para o agrupamento \emph{j} computamos $p_{ij}$, a probabilidade que um membro do agrupamento \emph{i} pertence à classe \emph{j} como: $p_{ij} = \dfrac{m_{ij}}{m_{i}}$, onde $m_{i}$ é o número de objetos no agrupamento \emph{i} e $m_{ij}$ é o número de objetos da classe \emph{j} no agrupamento \emph{i}. Utilizando esta distribuição de classe, a entropia de cada agrupamento \emph{i} é calculada usando a fórmula padrão \cite{nath:Tan2006}, $e_{i} = -\sum_{j=1}^{L}p_{ij}log_{2}p_{ij}$, onde \emph{L} é o número de classes. A entropia total para um conjunto de agrupamentos é calculada como sendo o somatório das entropias de cada agrupamento pelo tamanho de cada agrupamento, ou seja, $e = \sum_{i=1}^{K}\dfrac{m_{i}}{m}e_{i}$, onde \emph{K} é o número de agrupamentos e \emph{m} é o número total de padrões de dados.

Assim, os aspectos observados ao final do treinamento e que definiram o desempenho das abordagens estudadas foram os seguintes: percentagem da entropia para o conjunto de teste e tempo computacional para o conjunto de teste.

\subsection{Método de comparação entre as abordagens}
O teste de hipótese \cite{nath:Mitchell1997} é um teste bastante usado quando as variáveis envolvidas estão sujeitas à variabilidade, ou seja, sofrem mudanças. O teste de hipótese chamado \textit{Teste T} é um teste estatístico, baseado na distribuição \textit{t} de \textit{Student}. O teste \textit{t} de \textit{Student} é um teste de hipótese paramétrico, sendo que as duas variáveis envolvidas são intervalares e ambas têm distribuição normal. Quando se pretende comparar dois grupos, a hipótese nula representa que a diferença das médias é zero, isto é, não há diferenças entre os grupos. Assim, esta hipótese determina se um dado resultado é, ou não, estatisticamente significativo.

\section{Experimentos}
%O problema de reconhecimento de padrões aborda a classificação de um conjunto de dados baseado em suas características. Este problema pode ser usado com várias técnicas, como: estatísticas, árvores de decisão e RNAs. Muitas vezes confundida com classificação, o agrupamento de dados envolve a descoberta de classes bem definidas nos dados de entrada. Neste caso, as classes não são conhecidas previamente e a solução do problema de agrupamento envolve o aprendizado não-supervisionado \cite{nath:Braga2000}.

Neste trabalho usamos as redes SOM e SOINN com o propósito de criar classificadores que po\-ssam ser usados como sistemas de descoberta de padrões aplicados ao problema de agrupamento. Para a avaliação dos resultados utilizamos a técnica estatística 10-\emph{folds cross-validation} com 30 execuções sobre cada método de agrupamento com uma medida de erro baseada na entropia.

\subsection{Resultados Experimentais}
Para todas as técnicas híbridas de agrupamento de dados (e.g. SOMK, SOMAK, ASCAK e SOINAK) foram realizados os
mesmos experimentos. Em todos estes experimentos os dados são normalizados no intervalo entre [-1, 1]. Em relação ao parâmetro \emph{m} dos métodos propostos deste trabalho, que representa o número de formigas do algoritmo AK, o valor de  m = 2 obteve melhores resultados em \cite{nath:Souza2009a} \cite{nath:Souza2009b} comparado com m = 4 sugerido por Marco Dorigo \cite{nath:Dorigo2000}. Os experimentos apresentam os resultados dos métodos para a obtenção dos as\-pec\-tos observados nas 30 execuções realizadas com os valores de mínimo, máximo, média e \textit{Std}. 

\subsubsection{Desempenho}
Nas Tabelas ~\ref{tab:ResultadosEntropiaDadosSinteticos} e ~\ref{tab:ResultadosEntropiaDadosReais} são apresentados os resultados das técnicas de a\-gru\-pa\-men\-tos para os dados sintéticos e reais respectivamente, intervalos de confiança (IC), usando a normalização como método de pré-processamento para obtenção da taxa de erro baseada na entropia. O parâmetro desvio padrão, \textit{Std} apresentou um valor menor para os algoritmos de agrupamento K-médias, SOM e SOINN respeitando esta ordem em sua grande maioria para os conjuntos de dados, enquanto que para os métodos de agrupamento baseados em dois estágios (SOMK, SOMAK, ASCAK e SOINAK) os valores dos desvios padrões são mais altos. As Tabelas ~\ref{tab:ResultadosEntropiaDadosSinteticos} e ~\ref{tab:ResultadosEntropiaDadosReais} apresentaram uma grande variabilidade no \textit{Std}, destacando uma desvantagem dos métodos de agrupamento em dois estágios.

\setlength{\tabcolsep}{0.95pt}
\begin{table}[h!]
\begin{small}
\begin{center}
  \caption{Resultados dos métodos com 30 execuções cada obtendo entropia dos dados sintéticos (conjunto de teste) com os IC.}
  \vspace{-10pt}
    \label{tab:ResultadosEntropiaDadosSinteticos}
    \begin{tabular} {l*{6}{c}r} \\ \hline
        Dados & Métodos & Mínimo & Máximo & Média & Desvio & \multicolumn{2}{l}{Int. de Confiança} \\
        \cline{7-8}
        & & & & & & Limite & Limite \\
        & & & & & & Inferior & Superior \\ \hline
        & SOM & 0.0712 & 0.2004 & 0.1304 & 0.0325 & 0.1187 & 0.1420 \\
        & K-médias & \textbf{0.0112} & \textbf{0.0289} & \textbf{0.0165} & \textbf{0.0035} & \textbf{0.0152} & \textbf{0.0177}\\
        & ASCA & 0.0754 & 0.2102 & 0.1132 & 0.0266 & 0.1036 & 0.1227\\
        \multirow{-5}{0.1cm}{
        \begin{sideways}\parbox{2.0cm}{\emph{Lines} (I)}\end{sideways}} & SOINN & 0.0578 & 0.1035 & 0.0813 & 0.0081 & 0.0784 & 0.0841 \\
        & SOMK & 0.4492 & 0.5042 & 0.4789 &	0.0130 & 0.4742 & 0.4835\\
        & SOMAK & 0.0589 & 0.2969 & 0.1887 & 0.0534 & 0.1695 & 0.2078\\
        & ASCAK & 0.0833 & 0.6869 & 0.3621 & 0.1624 & 0.3039 & 0.4202\\
        & SOINAK & 0.0353 & 0.4637 & 0.2722 & 0.1173 & 0.2302 & 0.3141\\ \hline
        & SOM & 0.0724 & 0.1541 & 0.1095 & 0.0216 & 0.1017 & 0.1172\\
        & K-médias & 0.3558 & 0.4342 & 0.4029 &	0.0203 & 0.3956 & 0.4101\\
        & ASCA & 0.1380 & 0.4860 & 0.3432 &	0.0845 & 0.3129 & 0.3734\\
        \multirow{-7}{0.1cm}{
        \begin{sideways}\parbox{3.0cm}{\emph{Banana} (II)}\end{sideways}} & SOINN & 0.0598 & \textbf{0.0843} & \textbf{0.0819} & \textbf{0.0049} & \textbf{0.0801} & \textbf{0.0836}\\
        & SOMK & 0.3257 & 0.4728 & 0.4253 &	0.0318 & 0.4139 & 0.4366\\
        & SOMAK & 0.1488 & 0.5069 & 0.3326 & 0.0877 & 0.3012 & 0.3639\\
        & ASCAK & 0.0696 & 0.4420 & 0.2310 & 0.1017 & 0.1946 & 0.2673\\
        & SOINAK & \textbf{0.0099} & 0.3614 & 0.1840 & 0.0835 & 0.1541 & 0.2138\\ \hline
        & SOM &	0.2502 & 0.3314 & 0.2831 & 0.0236 & 0.2746 & 0.2915\\
        & K-médias & 0.4551 & 0.5063 & 0.4909 &	\textbf{0.0110} & 0.4869 & 0.4948\\
        & ASCA & 0.1313 & 0.4986 & 0.3856 &	0.0880 & 0.3541 & 0.4170\\
        \multirow{-9}{0.1cm}{
        \begin{sideways}\parbox{4.0cm}{\emph{Highleyman} (III)}\end{sideways}} & SOINN & 0.0575 & \textbf{0.2220} & \textbf{0.0870} & 0.0259 & \textbf{0.0777} & \textbf{0.0962}\\
        & SOMK & 0.4044 & 0.4939 & 0.4539 & 0.0257 & 0.4447 & 0.4630\\
        & SOMAK & 0.2429 & 0.5207 & 0.4000 & 0.0720 & 0.3742 & 0.4257\\
        & ASCAK & 0.0541 & 0.3770 & 0.2215 & 0.0739 & 0.1950 & 0.2479\\
        & SOINAK & \textbf{0.0291} & 0.2780 & 0.1504 & 0.0688 & 0.1257 & 0.1750\\ \hline
        & SOM & 0.3882 & 0.4852 & 0.4481 & 0.0203 & 0.4408 & 0.4553\\
        & K-médias & 0.5052 & 0.5228 & 0.5145 &	\textbf{0.0038} & 0.5131 & 0.5158\\
        & ASCA & 0.2066 & 0.5096 & 0.4159 & 0.0763 & 0.3885 & 0.4432\\
        \multirow{-7}{0.1cm}{
        \begin{sideways}\parbox{3.0cm}{\emph{Spherical} (IV)}\end{sideways}} & SOINN & \textbf{0.0392} & \textbf{0.2224} & \textbf{0.0838} & 0.0280 & \textbf{0.0737} & \textbf{0.0938}\\
        & SOMK & 0.5042 & 0.5296 & 0.5208 & 0.0069 & 0.5183 & 0.5232\\
        & SOMAK & 0.2644 & 0.5131 & 0.4338 & 0.0627 & 0.4113 & 0.4562\\
        & ASCAK & 0.0641 & 0.3611 & 0.1947 & 0.0805 & 0.1658 & 0.2235\\
        & SOINAK & 0.0465 & 0.3710 & 0.2185 & 0.0790 & 0.1902 & 0.2467\\ \hline
        & SOM & 0.3905 & 0.4755 & 0.4343 & 0.0230 & 0.4260 & 0.4425\\
        & K-médias & 0.3726 & 0.4575 & 0.4207 &	\textbf{0.0198} & 0.4136 & 0.4277\\
        & ASCA & 0.0906 & 0.4825 & 0.3766 &	0.0864 & 0.3456 & 0.4075\\
        \multirow{-8}{0.1cm}{
        \begin{sideways}\parbox{4.0cm}{\emph{Simple Problem} (V)}\end{sideways}} & SOINN & \textbf{0.0364} & \textbf{0.2224} & \textbf{0.0829} & 0.0287 & \textbf{0.0726} & \textbf{0.0931}\\
        & SOMK & 0.4515 & 0.5286 & 0.4934 &	0.0223 & 0.4854 & 0.5013\\
        & SOMAK & 0.0376 & 0.5039 & 0.3267 & 0.1103 & 0.2872 & 0.3661\\
        & ASCAK & 0.1112 & 0.5278 & 0.2116 & 0.0907 & 0.1791 & 0.2440\\
        & SOINAK & 0.0181 & 0.3259 & 0.1686 & 0.0898 & 0.1364 & 0.2007\\ \hline
  \end{tabular}
\end{center}
\end{small}
\end{table}

\setlength{\tabcolsep}{0.95pt}
\begin{table}[h!]
\begin{small}
\begin{center}
  \caption{Resultados dos métodos com 30 execuções cada obtendo entropia dos dados reais (conjunto de teste) com os IC.}
   \vspace{-10pt}
    \label{tab:ResultadosEntropiaDadosReais}
    \begin{tabular} {l*{6}{c}r} \\ \hline
        Dados & Métodos & Mínimo & Máximo & Média & Desvio & \multicolumn{2}{l}{Int. de Confiança} \\
        \cline{7-8}
        & & & & & & Limite & Limite \\
        & & & & & & Inferior & Superior \\ \hline
        & SOM & 0.2955 & 0.3491 & 0.3224 & \textbf{0.0112} & 0.3183 & 0.3264\\
        & K-médias & 0.0685 & 0.1869 & 0.1052 &	0.0321 & 0.0937 & 0.1166\\
        & ASCA & 0.1064 & 0.2661 & 0.2001 & 0.0403 & 0.1856 & 0.2145\\
        \multirow{-8}{0.1cm}{
        \begin{sideways}\parbox{4.0cm}{\emph{Balance Scale} (VI)}\end{sideways}} & SOINN & 0.0593 & \textbf{0.1421} & \textbf{0.0839} & 0.0179 & \textbf{0.0774} & \textbf{0.0903}\\
        & SOMK & 0.3232 & 0.4425 & 0.3784 &	0.0297 & 0.3677 & 0.3890\\
        & SOMAK & 0.0705 & 0.3761 & 0.2093 & 0.0692 & 0.1845 & 0.2340\\
        & ASCAK & 0.0833 & 0.4816 & 0.2606 & 0.0901 & 0.2283 & 0.2928\\
        & SOINAK & \textbf{0.0384} & 0.3503 & 0.1896 & 0.0875 & 0.1582 & 0.2209\\ \hline
        & SOM &	0.4453 & 0.4929 & 0.4743 & \textbf{0.0106} & 0.4705 & 0.4780\\
        & K-médias & 0.0685 & 0.1870 & 0.1145 & 0.0334 & 0.1025 & 0.1264\\
        & ASCA & 0.1872 & 0.4120 & 0.2954 & 0.0692 & 0.2706 & 0.3201\\
        \multirow{-7}{0.1cm}{
        \begin{sideways}\parbox{3.0cm}{\emph{CMC} (VII)}\end{sideways}} & SOINN & \textbf{0.0104} & \textbf{0.1039} & \textbf{0.0703} &	0.0191 & \textbf{0.0634} & \textbf{0.0771}\\
        & SOMK & 0.4025 & 0.4821 & 0.4414 & 0.0201 & 0.4342 & 0.4485\\
        & SOMAK & 0.1432 & 0.4182 & 0.2546 & 0.0745 & 0.2279 & 0.2812\\
        & ASCAK & 0.0481 & 0.4274 & 0.2245 & 0.0869 & 0.1934 & 0.2555\\
        & SOINAK & 0.0170 & 0.4189 & 0.2103 & 0.1010 & 0.1741 & 0.2464\\ \hline
        & SOM &	0.3212 & 0.4324 & 0.3723 & 0.0289 & 0.3619 & 0.3826\\
        & K-médias & 0.0722 & 0.1840 & 0.1086 &	0.0262 & 0.0992 & 0.1179\\
        & ASCA & 0.0832 & 0.2328 & 0.1622 &	0.0454 & 0.1459 & 0.1784\\
        \multirow{-8}{0.1cm}{
        \begin{sideways}\parbox{4.0cm}{\emph{Dermatology} (VIII)}\end{sideways}} & SOINN & \textbf{0.0441} & \textbf{0.1597} & \textbf{0.0804} & 0.0225 & \textbf{0.0723} & \textbf{0.0884}\\
        & SOMK & 0.3170 & 0.4319 & 0.3676 & 0.0246 & 0.3587 & 0.3764\\
        & SOMAK & 0.1420 & 0.3656 & 0.2381 & 0.0604 & 0.2164 & 0.2597\\
        & ASCAK & 0.0925 & 0.4028 & 0.2220 & 0.0806 & 0.1931 & 0.2508\\
        & SOINAK & 0.0759 & 0.4940 & 0.2387 & 0.0973 & 0.2038 & 0.2735\\ \hline
        & SOM & 0.4738 & 0.5079 & 0.4914 & 0.0078 & 0.4886 & 0.4941\\
        & K-médias & 0.0700 & \textbf{0.1910} & 0.1157 & 0.0379 & 0.1021 & 0.1292\\
        & ASCA & 0.1773 & 0.4988 & 0.3643 &	0.0866 & 0.3333 & 0.3952\\
        \multirow{-7}{0.1cm}{
        \begin{sideways}\parbox{3.0cm}{\emph{Diabetes} (IX)}\end{sideways}} & SOINN	& 0.0569 & 0.2133 & \textbf{0.0892} & 0.0314 & \textbf{0.0779} &\textbf{0.1004}\\
        & SOMK & 0.5011 & 0.5250 & 0.5144 & \textbf{0.0062} & 0.5121 & 0.5166\\
        & SOMAK & 0.1933 & 0.5282 & 0.4643 & 0.0739 & 0.4378 & 0.4907\\
        & ASCAK & 0.1241 & 0.3539 & 0.2393 & 0.0606 & 0.2176 & 0.2609\\
        & SOINAK & \textbf{0.0161} & 0.3778 & 0.1999 & 0.0849 & 0.1695 & 0.2302\\ \hline
        & SOM & 0.3635 & 0.4451 & 0.4080 & \textbf{0.0197} & 0.4009 & 0.4150\\
        & K-médias & 0.0748 & 0.2021 & 0.1310 &	0.0364 & 0.1179 & 0.1440\\
        & ASCA & 0.0952 & 0.3292 & 0.2149 &	0.0593 & 0.1936 & 0.2361\\
        \multirow{-7}{0.1cm}{
        \begin{sideways}\parbox{3.0cm}{\emph{Glass} (X)}\end{sideways}} & SOINN & \textbf{0.0369} & \textbf{0.1907} & \textbf{0.0998} &	0.0431 & \textbf{0.0843} & \textbf{0.1152}\\
        & SOMK & 0.3286 & 0.4338 & 0.3758 &	0.0281 & 0.3657 & 0.3858\\
        & SOMAK & 0.0868 & 0.3461 & 0.2331 & 0.0624 & 0.2107 & 0.2554\\
        & ASCAK	& 0.1174 & 0.4268 & 0.2467 & 0.0736 & 0.2203 & 0.2730\\
        & SOINAK & 0.0872 & 0.3804 & 0.2400 & 0.0833 & 0.2101 & 0.2698\\ \hline
  \end{tabular}
\end{center}
\end{small}
\end{table}

SOMK apresentou neste trabalho, algumas deficiências, como: o desempenho, onde a medida de entropia tem sido maior que os demais métodos propostos. Na grande maioria dos dados o método proposto SOINAK mostrou uma medida de entropia menor nos respectivos parâmetros de mínimo, máximo e média, quando comparado as técnicas de agrupamento SOM, K-médias, ASCA, SOINN, SOMK, SOMAK e ASCAK. Portanto, a técnica híbrida de agrupamento SOINAK apresenta melhor eficiência em comparação com SOMK, SOMAK e ASCAK, que são também métodos compostos em dois estágios.

\subsubsection{Tempo Computacional}
Na Tabela \ref{tab:ResultadosTempoComputacional} são apresentados os tempos computacionais para os métodos usados neste trabalho para os dados sintéticos e reais. O algoritmo K-médias tem sido, em todos os dados (sintéticos e reais), o método de agrupamento mais rápido computacionalmente, pois este é um algoritmo simples, de uma única fase. No entanto, o algoritmo k-médias apresentou uma medida de entropia maior mostrado nas Tabelas \ref{tab:ResultadosEntropiaDadosSinteticos} e \ref{tab:ResultadosEntropiaDadosReais}, quando comparado às demais técnicas de agrupamentos de dados. Em seguida, foi observado que a técnica híbrida SOINAK apresentou resultados significativos, ou seja, melhor em comparação com os métodos de agrupamento SOMK, SOMAK e ASCAK, porém o SOINAK obteve o tempo computacional duas vezes maior que os métodos SOMK e SOMAK como podem ser vistos na Tabela \ref{tab:ResultadosTempoComputacional}.

\setlength{\tabcolsep}{0.95pt}
\begin{table}[h!]
\begin{small}
\begin{center}
  \caption{Resultados dos métodos com 30 execuções obtendo tempo computacional dos dados sintéticos e reais (dados de teste).}
  \vspace{-10pt}
  \label{tab:ResultadosTempoComputacional}
    \begin{tabular} {l*{12}{c}r} \\ \hline
        Dados & Métodos & Mínimo & Máximo & Média & Desvio & ~~~ & Dados & Métodos & Mínimo & Máximo & Média & Desvio \\ \hline
        & SOM & 5.467 & 6.189 & 5.7178 & 0.1606 & & & SOM & 3.3382 & 3.6947 & 3.4666 & 0.0763 \\
        & K-médias & \textbf{0.4372} & \textbf{0.5228} & \textbf{0.4462} & \textbf{0.0151} & & & K-médias & \textbf{0.0560} & \textbf{0.0863} & \textbf{0.0655} & \textbf{0.0077} \\
        & ASCA & 11.676 & 14.922 & 13.098 &	0.7821 & & & ASCA & 3.3194 & 6.2293 & 4.7890 & 0.7734 \\
        \multirow{-5}{0.1cm}{
        \begin{sideways}\parbox{2.0cm}{\emph{Lines} (I)}\end{sideways}} & SOINN & 14.270 & 18.089 & 16.791 & 1.1233 & ~~~ & \multirow{-8}{0.1cm}{
        \begin{sideways}\parbox{4.0cm}{\emph{Balance Scale} (VI)}\end{sideways}} & SOINN & 5.0158 & 6.5270 & 5.8638 & 0.3422 \\
        & SOMK & 6.6046 & 7.3494 & 6.8666 &	0.1643 & & & SOMK & 4.3049 & 4.6706 & 4.4508 & 0.0840 \\
        & SOMAK & 6.3644 & 7.1004 & 6.6946 & 0.1844 & & & SOMAK & 4.2297 & 4.9395 & 4.4614 & 0.1559 \\
        & ASCAK & 11.680 & 14.928 & 13.104 & 0.7832 & & & ASCAK & 3.8090 & 7.0879 & 5.1767 & 1.0192 \\
        & SOINAK & 14.882 & 19.187 & 17.401 & 1.1320 & & & SOINAK & 5.4872 & 7.7862 & 6.3723 & 0.5532 \\ \hline
        & SOM & 6.2183 & 6.8441 & 6.4857 & 0.1378 & & & SOM & 13.528 & 14.845 & 14.043 & 0.3043 \\ 
        & K-médias & \textbf{0.3366} & \textbf{0.3571} & \textbf{0.3448} & \textbf{0.0051} & & & K-médias & \textbf{0.0540} & \textbf{0.2537} & \textbf{0.0656} & \textbf{0.0358} \\
        & ASCA & 13.006 & 26.586 & 20.090 & 2.9588 & & & ASCA & 41.078 & 238.57 & 93.974 & 46.674 \\
        \multirow{-7}{0.1cm}{
        \begin{sideways}\parbox{3.0cm}{\emph{Banana} (II)}\end{sideways}} & SOINN & 15.375 & 17.653 & 16.575 & 0.5266 & ~~~ & \multirow{-7}{0.1cm}{
        \begin{sideways}\parbox{3.0cm}{\emph{CMC} (VII)}\end{sideways}} & SOINN	& 38.335 & 41.391 & 40.522 & 0.6491 \\
        & SOMK & 7.3520 & 8.2189 & 7.6472 &	0.1618 & & & SOMK & 14.909 & 16.240 & 15.434 & 0.3070 \\
        & SOMAK & 7.5242 & 8.5392 & 8.0676 & 0.2543 & & & SOMAK & 14.837 & 16.187 & 15.453 & 0.3266 \\
        & ASCAK & 13.009 & 26.703 & 20.891 & 3.0574 & & & ASCAK	& 44.166 & 238.88 & 101.97 & 49.371 \\
        & SOINAK & 15.841 & 19.068 & 17.398 & 0.8336 & & & SOINAK & 39.158 & 44.405 & 41.518 & 1.1583 \\ \hline
        & SOM &	6.2533 & 6.8994 & 6.4471 & 0.1222 & & & SOM & 28.938 & 29.297 & 29.087 & 0.0871 \\
        & K-médias & \textbf{0.3198} & \textbf{0.3347} & \textbf{0.3257} & \textbf{0.0039} & & & K-médias & \textbf{0.0436} & \textbf{0.1451} & \textbf{0.0607} & \textbf{0.0285} \\
        & ASCA & 10.869 & 23.804 & 18.036 &	3.7306 & & & ASCA & 0.7029 & 1.3905 & 1.0656 & 0.1643 \\
        \multirow{-9}{0.1cm}{
        \begin{sideways}\parbox{4.0cm}{\emph{Highleyman} (III)}\end{sideways}} & SOINN & 15.634 & 17.278 & 16.413 & 0.4788 & ~~~ & \multirow{-8}{0.1cm}{
        \begin{sideways}\parbox{4.0cm}{\emph{Dermatology} (VIII)}\end{sideways}} & SOINN & 1.5755 & 2.3479 & 2.0489 & 0.1694 \\
        & SOMK & 7.4056 & 8.0478 & 7.6023 &	0.1212 & & & SOMK & 29.943 & 30.304 & 30.102 & 0.1043 \\
        & SOMAK & 7.1529 & 8.3231 & 7.4052 & 0.2235 & & & SOMAK & 29.364 & 29.689 & 29.540 & 0.0855 \\
        & ASCAK & 12.706 & 24.795 & 19.618 & 3.4978 & & & ASCAK & 0.9596 & 1.8389 & 1.3897 & 0.2246 \\
        & SOINAK & 15.825 & 18.659 & 17.0861 & 0.7197 & & & SOINAK & 1.1560 & 2.6256 & 2.0888 & 0.3026 \\ \hline
        & SOM & 6.3861 & 6.8165 & 6.5760 & 0.1165 & & & SOM & 5.8072 & 6.2175 & 5.9322 & 0.0719 \\
        & K-médias & \textbf{0.3173} & \textbf{0.4190} & \textbf{0.3320} & \textbf{0.0263} & & & K-médias & \textbf{0.0394} & \textbf{0.2944} & \textbf{0.0546} & \textbf{0.0456} \\
        & ASCA & 7.2653 & 37.777 & 21.107 &	7.7648 & & & ASCA & 7.8359 & 44.228 & 21.904 & 8.5322 \\
        \multirow{-7}{0.1cm}{
        \begin{sideways}\parbox{3.0cm}{\emph{Spherical} (IV)}\end{sideways}} & SOINN & 14.065 & 18.406 & 16.874 & 1.2014 & ~~~ & \multirow{-7}{0.1cm}{
        \begin{sideways}\parbox{3.0cm}{\emph{Diabetes} (IX)}\end{sideways}} & SOINN	& 9.6132 & 10.863 & 10.092 & 0.2528 \\
        & SOMK & 7.5340 & 7.9477 & 7.7284 &	0.1193 & & & SOMK & 6.9667 & 7.3696 & 7.1129 & 0.0858 \\
        & SOMAK	& 7.5691 & 8.8254 & 8.2795 & 0.3354 & & & SOMAK & 6.5917 & 7.0380 & 6.7562 & 0.0958 \\
        & ASCAK	& 7.5509 & 39.579 & 26.424 & 9.4688 & & & ASCAK & 10.413 & 44.264 & 28.077 & 10.386 \\
        & SOINAK & 15.329 & 19.453 & 17.336 & 1.2029 & & & SOINAK & 9.8258 & 18.779 & 11.562 & 2.8513 \\ \hline
        & SOM & 6.2029 & 6.5876 & 6.3902 & 0.1037 & & & SOM & 4.4001 & 4.4820 & 4.4260 & 0.0192 \\
        & K-médias & \textbf{0.3148} & \textbf{0.3345} & \textbf{0.3194} & \textbf{0.0039} & & & K-médias & \textbf{0.0347} & \textbf{0.1030} & \textbf{0.0436} & \textbf{0.0158} \\
        & ASCA & 9.8263 & 27.929 & 16.159 & 1.0123 & & & ASCA & 0.6235 & 1.3029 & 1.0200 & 0.2019 \\
        \multirow{-8}{0.1cm}{
        \begin{sideways}\parbox{4.0cm}{\emph{Simple Problem} (V)}\end{sideways}} & SOINN & 7.3384 & 7.7445 & 7.5460 & 0.1089 & ~~~ & \multirow{-7}{0.1cm}{
        \begin{sideways}\parbox{3.0cm}{\emph{Glass} (X)}\end{sideways}} & SOINN & 0.3504 & 0.6346 & 0.4515 & 0.0617 \\
        & SOMK & 7.3409 & 8.8399 & 8.1355 &	0.3452 & & & SOMK & 5.1649 & 5.2485 & 5.1991 & 0.0256 \\
        & SOMAK & 7.3409 & 8.8399 & 8.1355 & 0.3452 & & & SOMAK & 4.6482 & 4.7988 & 4.7216 & 0.0386 \\
        & ASCAK & 8.3152 & 31.024 & 22.193 & 4.9208 & & & ASCAK & 0.7883 & 1.3803 & 1.1629 & 0.1440 \\
        & SOINAK & 14.847 & 18.495 & 16.746 & 0.9966 & & & SOINAK & 0.3490 & 0.8239 & 0.5406 & 0.1045 \\ \hline
  \end{tabular}
\end{center}
\end{small}
\end{table}

\subsubsection{Intervalo de Confiança}
\begin{figure*}
\centering
\subfigure[]{
\epsfig{figure=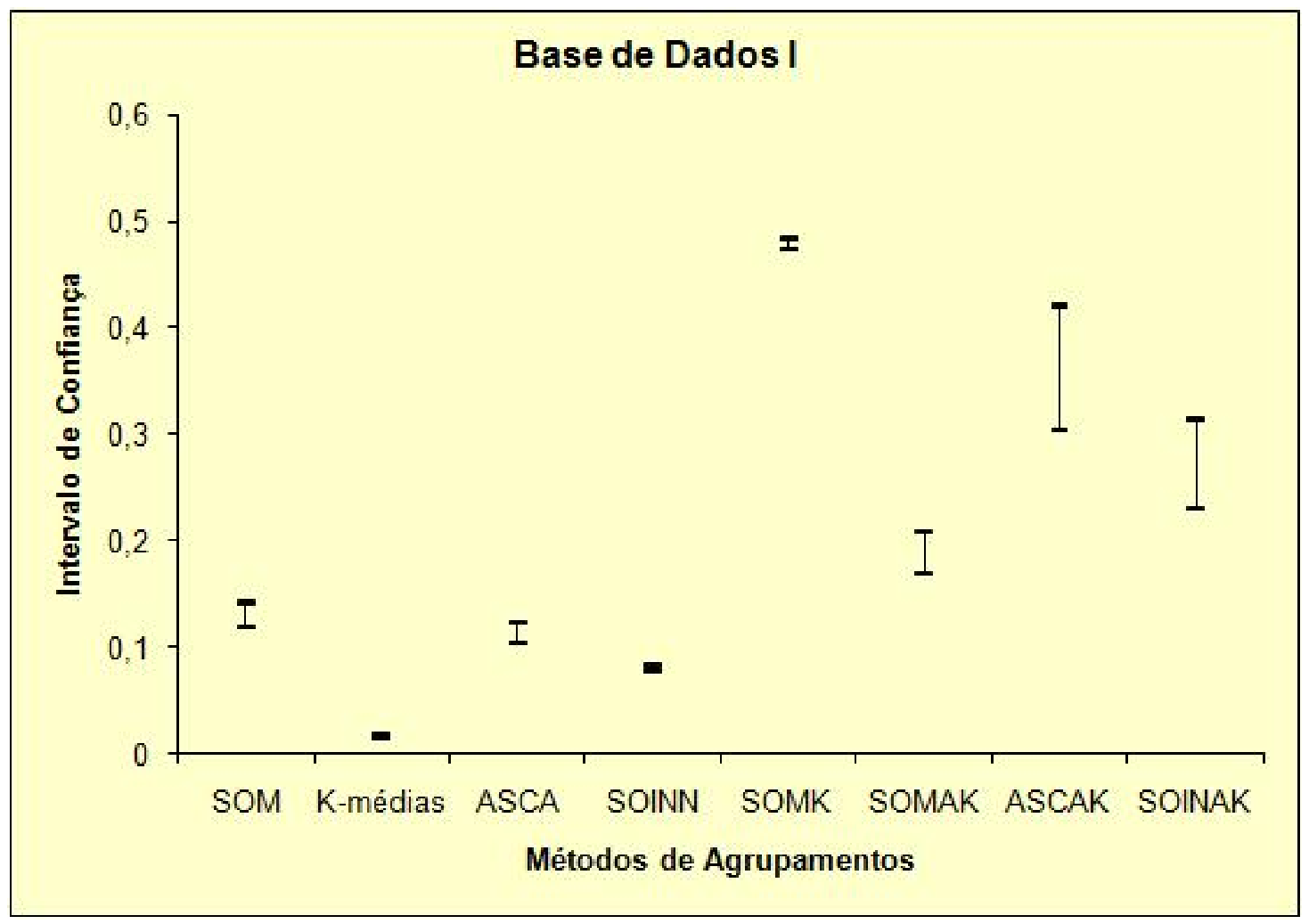,width=56mm}
}
\subfigure[]{
\epsfig{figure=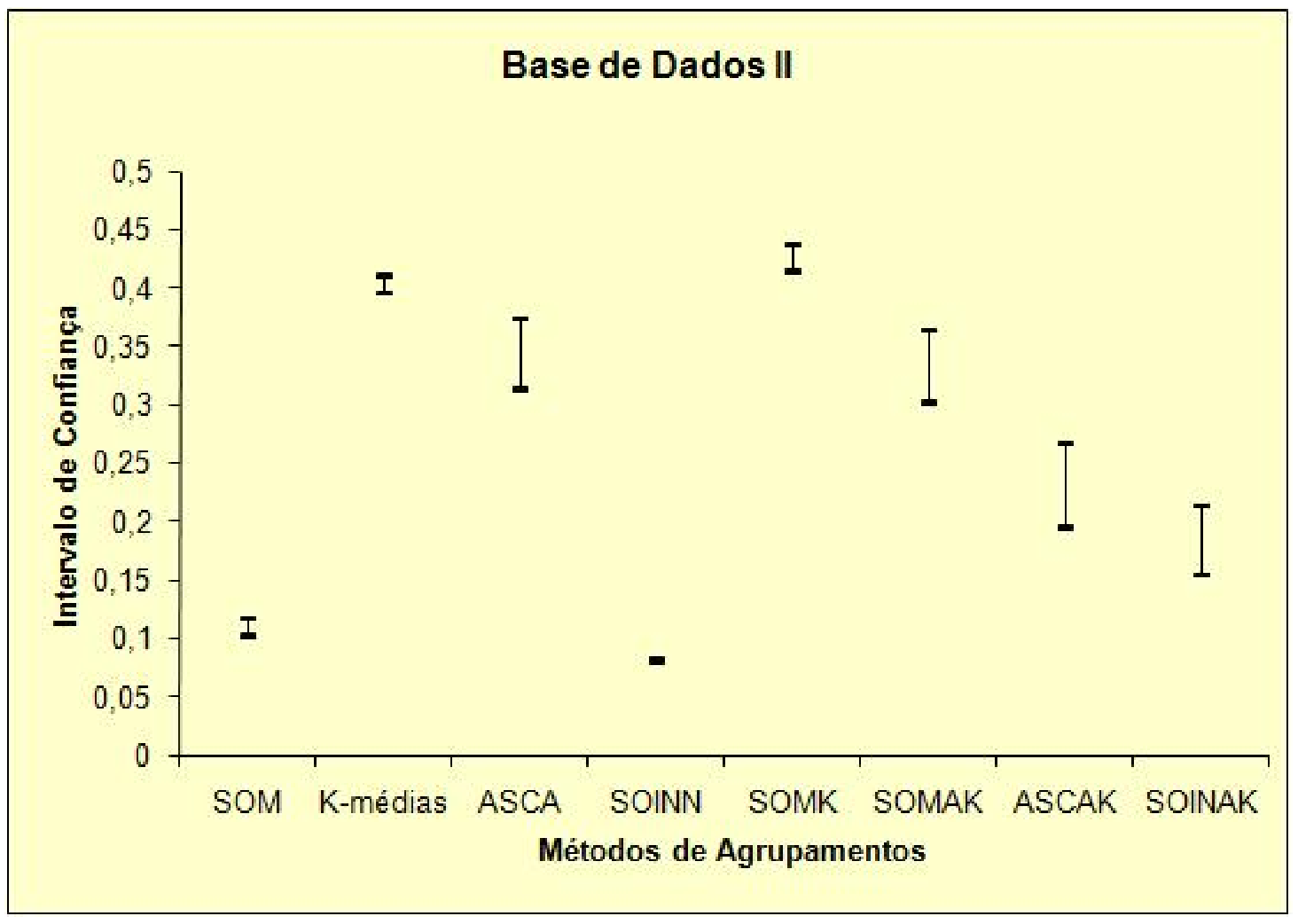,width=56mm}
}
\vspace{-14pt}
\caption{Intervalos de confiança da base \emph{Lines} (a) e \emph{Banana} (b).}
\label{fig:GraficoIntervaloConfiancaI}
\end{figure*}

\begin{figure*}
\centering
\subfigure[]{
\epsfig{figure=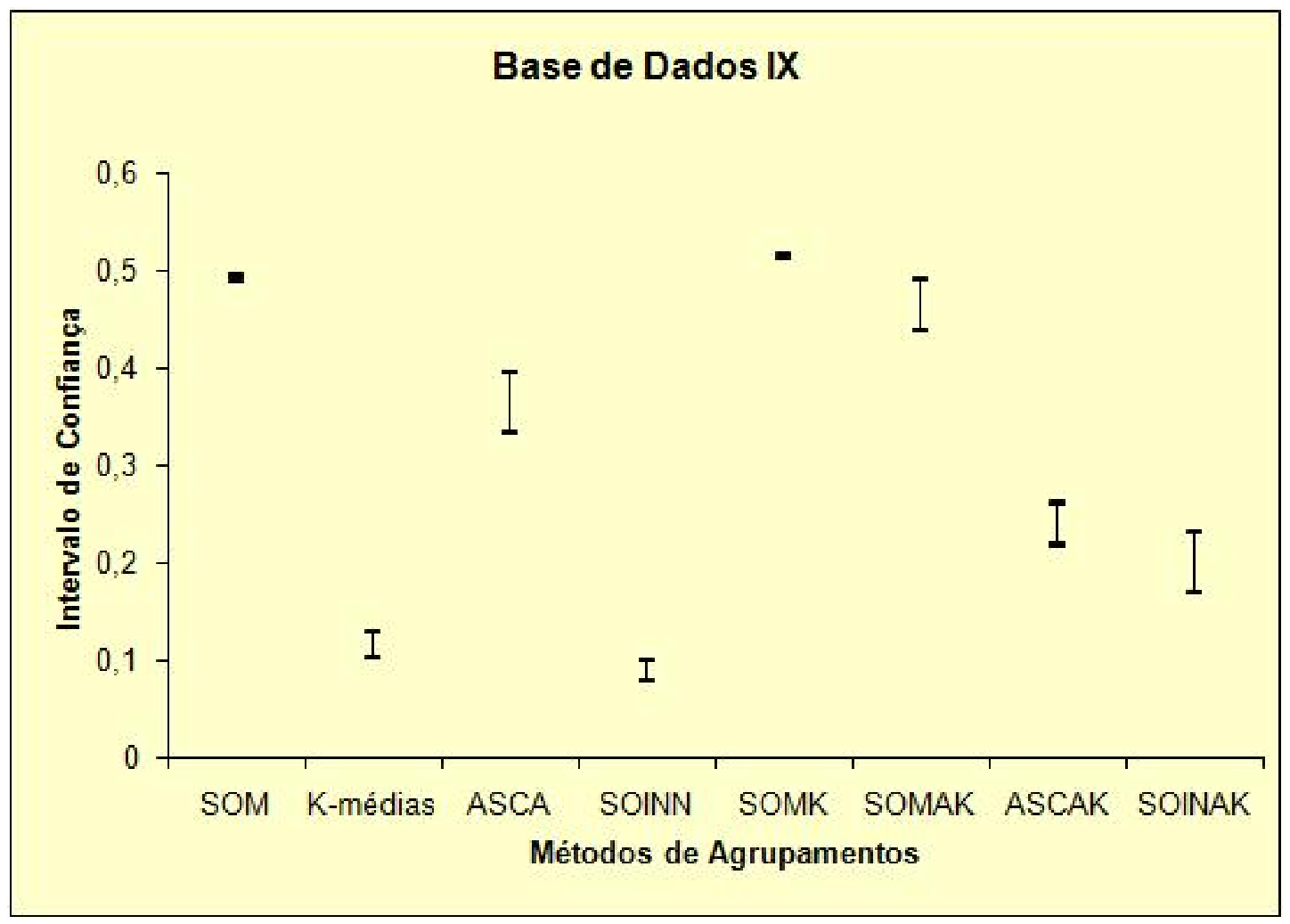,width=56mm}
}
\subfigure[]{
\epsfig{figure=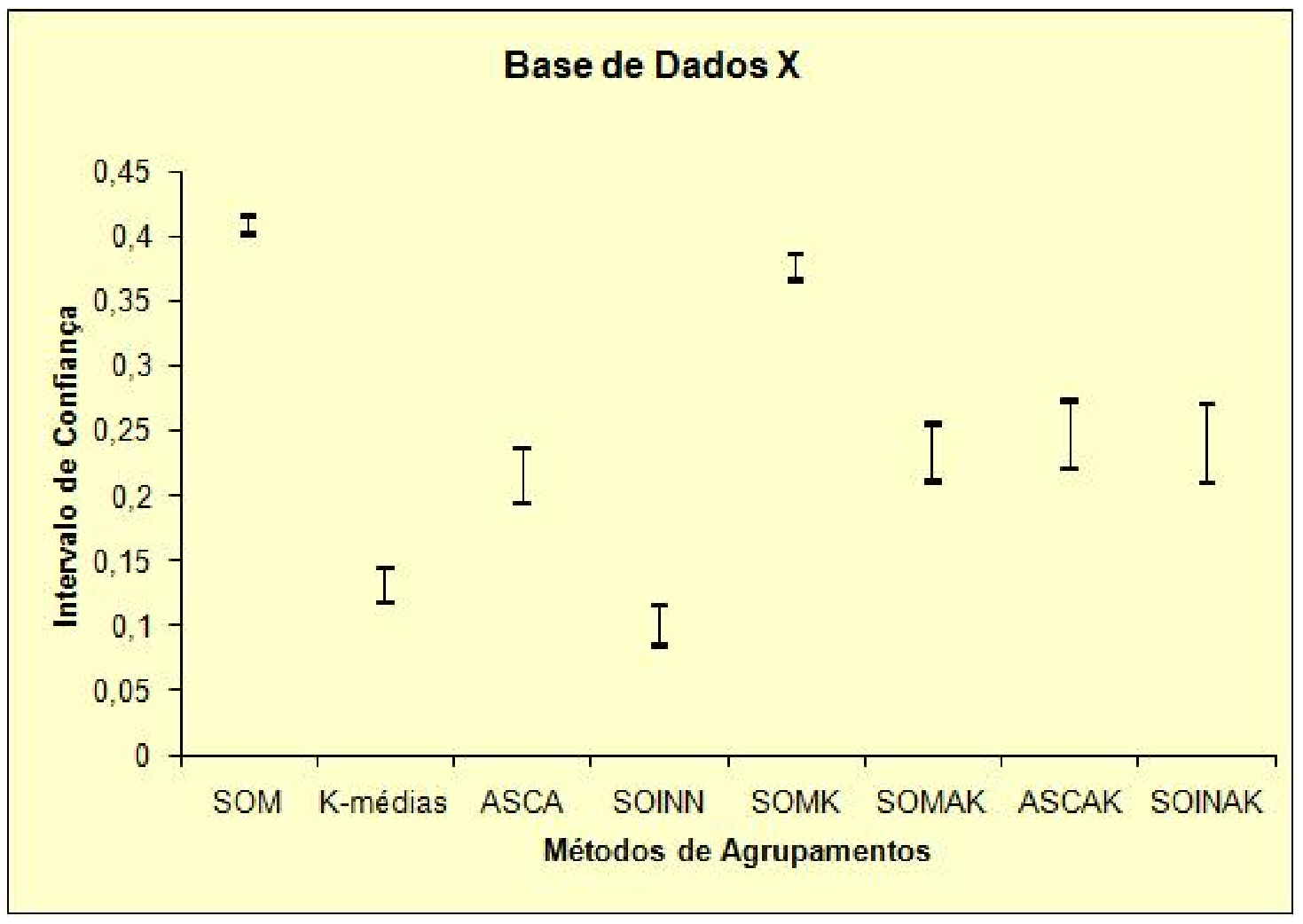,width=56mm}
}
\vspace{-14pt}
\caption{Intervalos de confiança da base \emph{Diabetes} (a) e \emph{Glass} (b).}
\label{fig:GraficoIntervaloConfiancaV}
\end{figure*}

As Tabelas \ref{tab:ResultadosEntropiaDadosSinteticos} e \ref{tab:ResultadosEntropiaDadosReais} apresentam os resultados dos IC obtidos das técnicas de agrupamentos para todos os dados, usando a média e \textit{Std} como base. Os Gráficos podem ser construídos a partir dos IC e as Figuras \ref{fig:GraficoIntervaloConfiancaI} a \ref{fig:GraficoIntervaloConfiancaV} mostram os IC para cada base de dados. 

\begin{figure*}
\centering
\subfigure[]{
\epsfig{figure=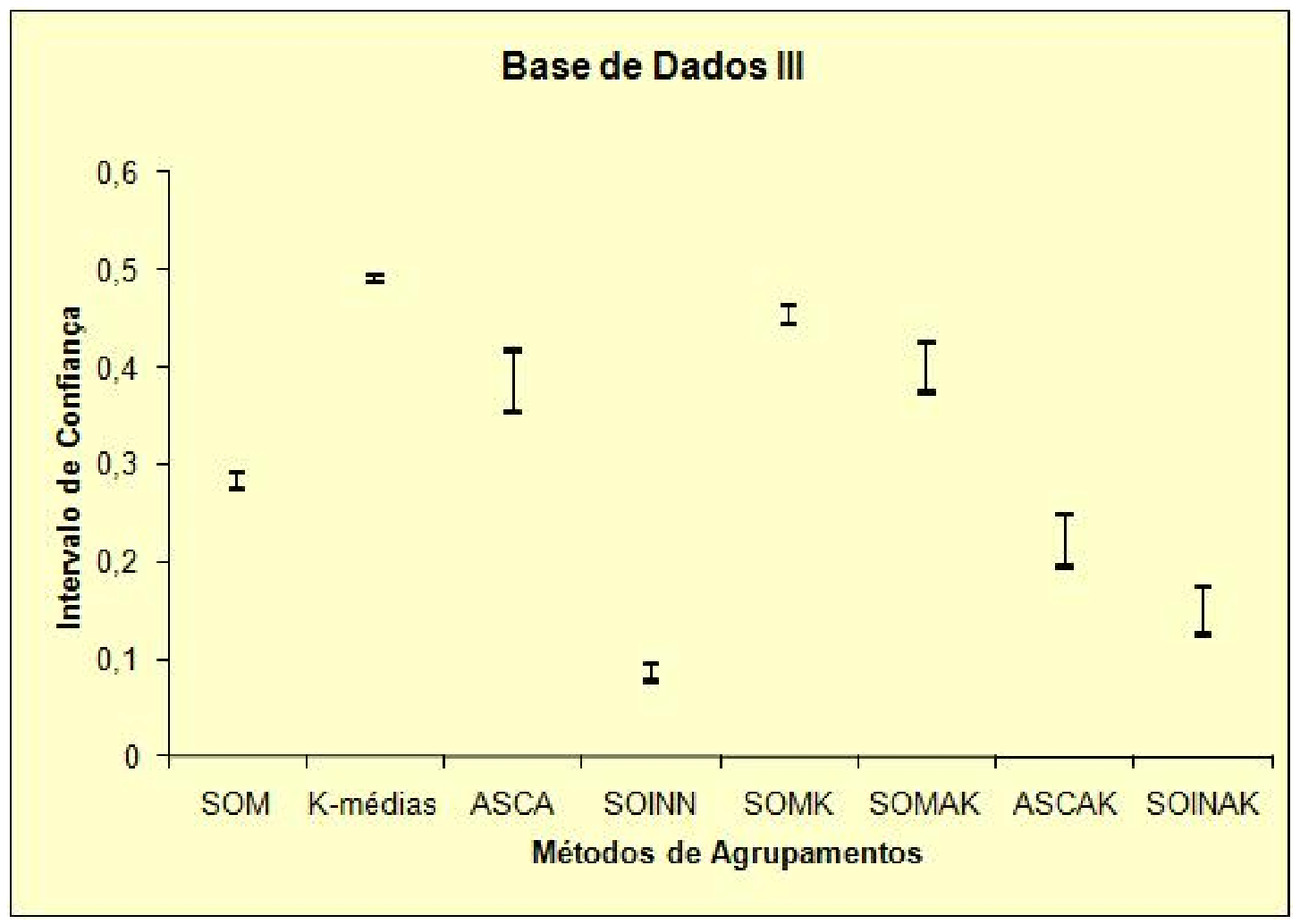,width=56mm}
}
\subfigure[]{
\epsfig{figure=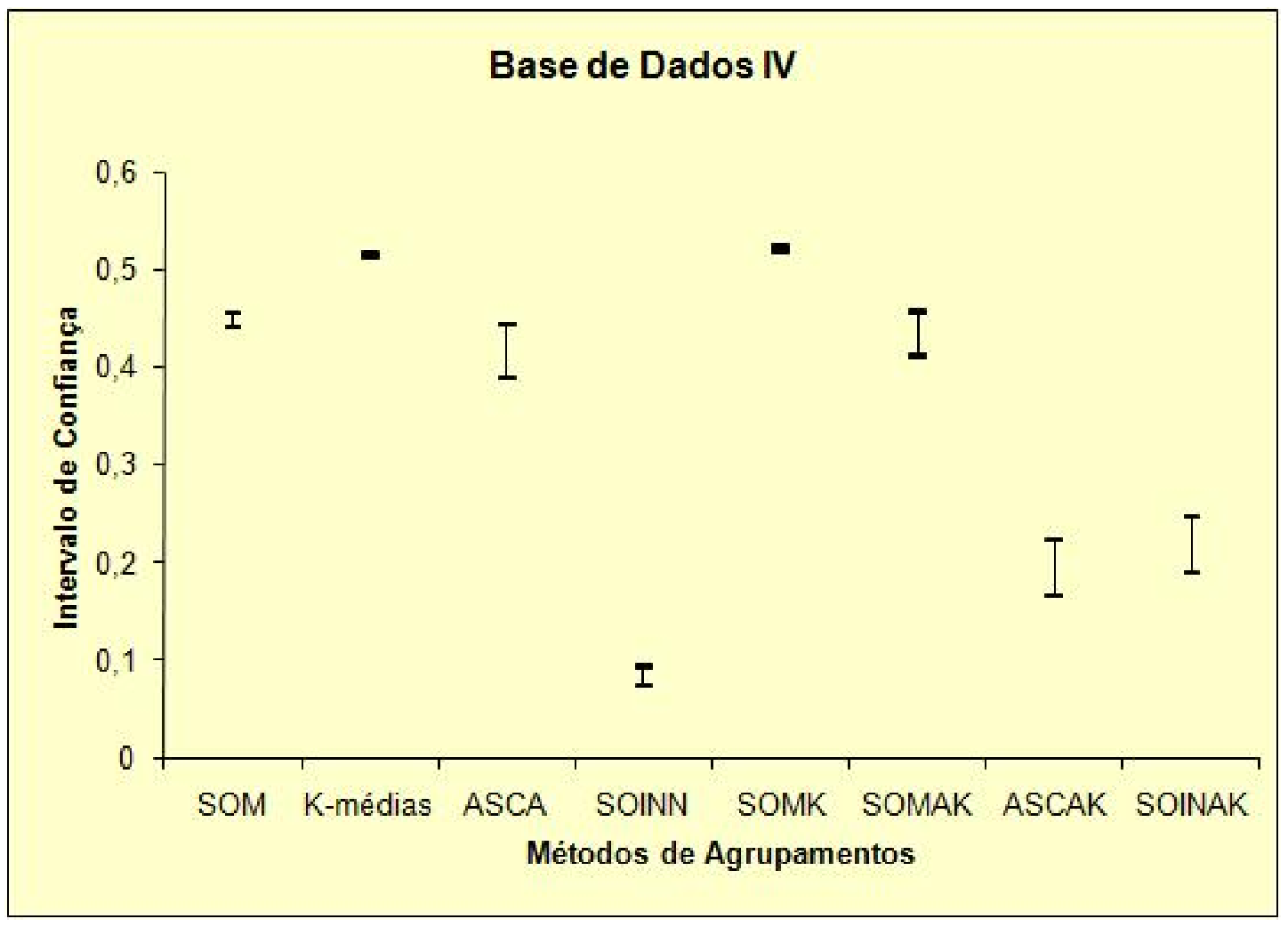,width=56mm}
}
\vspace{-14pt}
\caption{Intervalos de confiança da base \emph{Highleyman} (a) e \emph{Spherical} (b).}
\label{fig:GraficoIntervaloConfiancaII}
\end{figure*}

\begin{figure*}
\centering
\subfigure[]{
\epsfig{figure=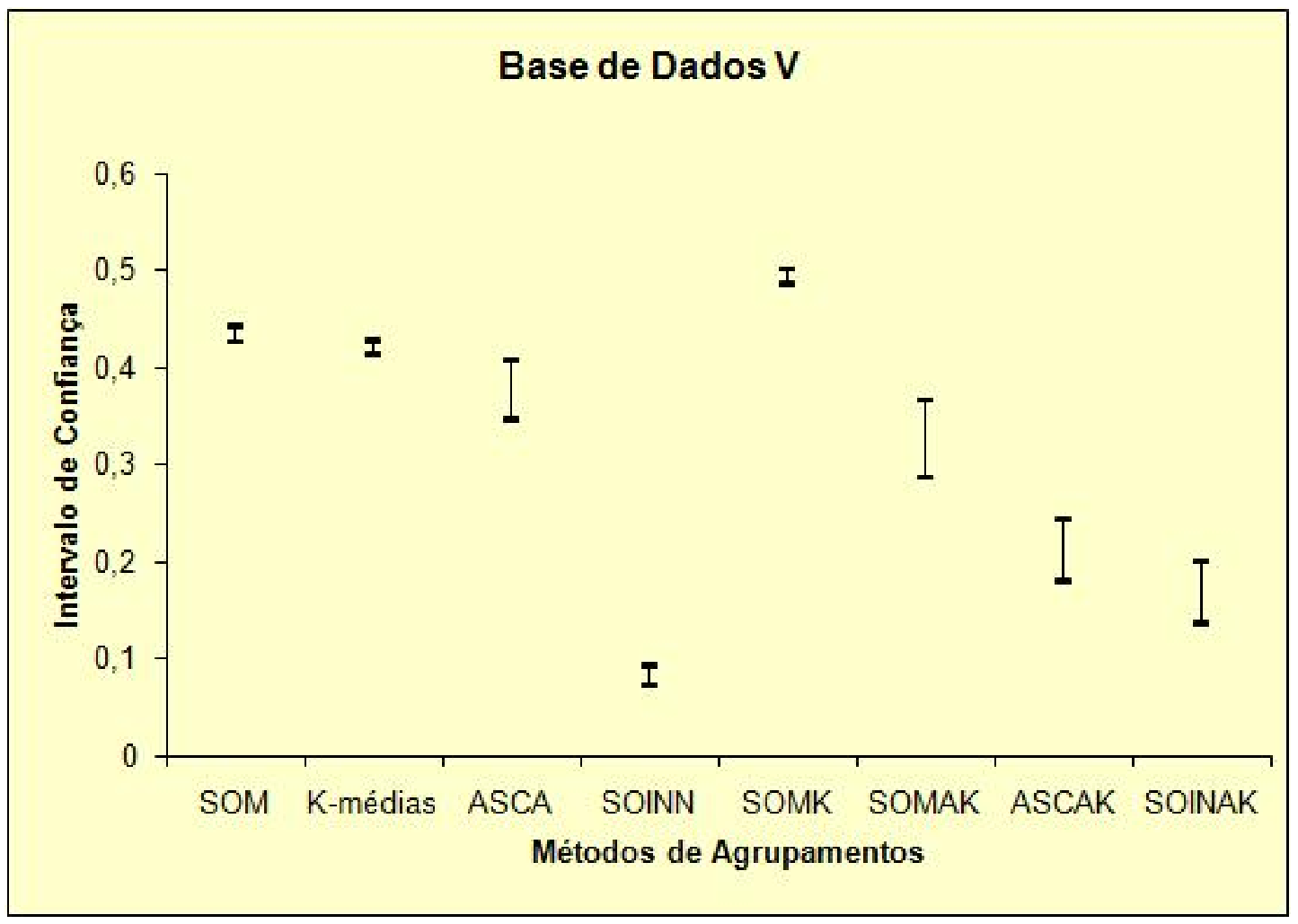,width=56mm}
}
\subfigure[]{
\epsfig{figure=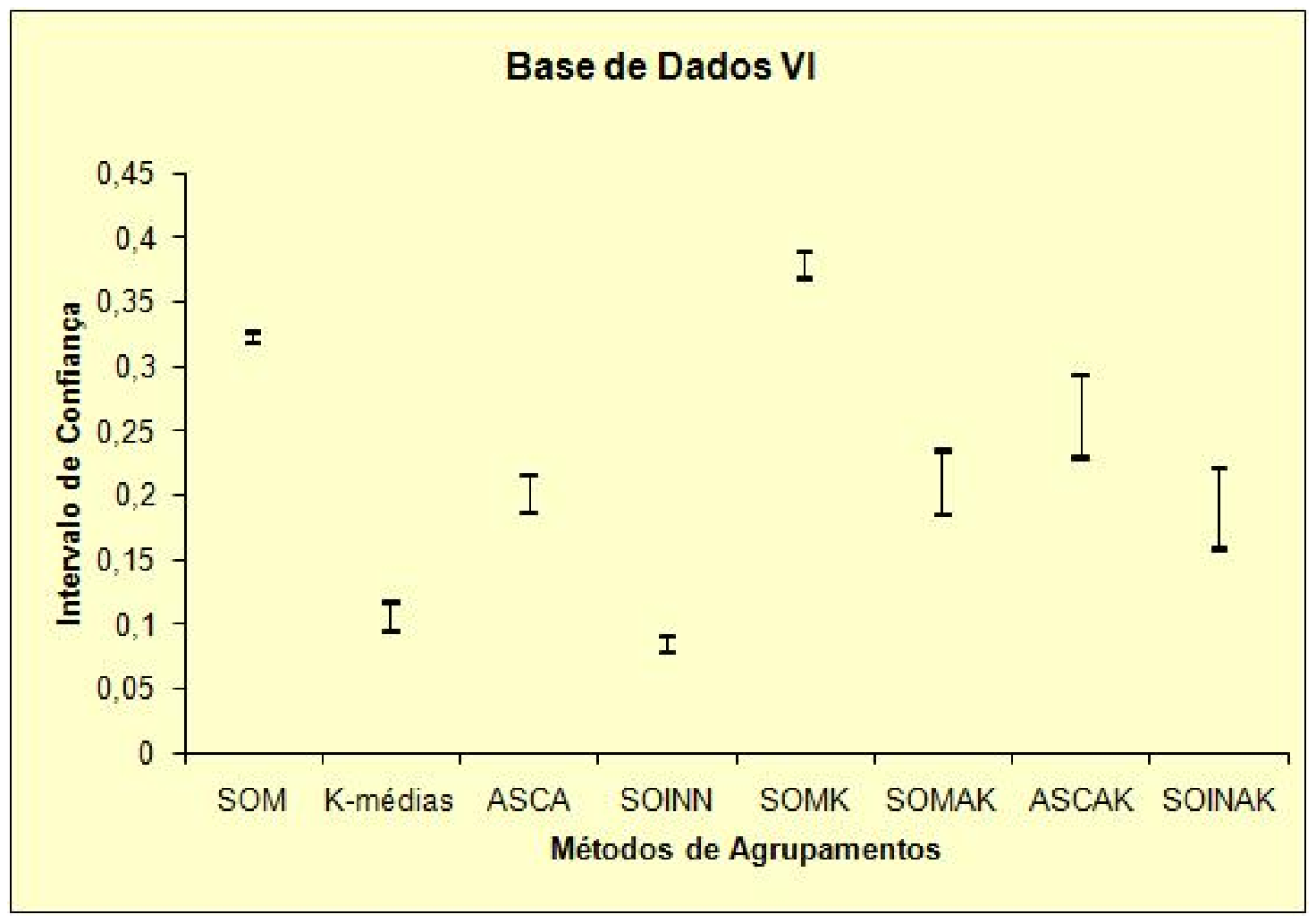,width=56mm}
}
\vspace{-14pt}
\caption{Intervalos de confiança da base \emph{Simple Problem} (a) e \emph{Balance Scale} (b).}
\label{fig:GraficoIntervaloConfiancaIII}
\end{figure*}

\begin{figure*}
\centering
\subfigure[]{
\epsfig{figure=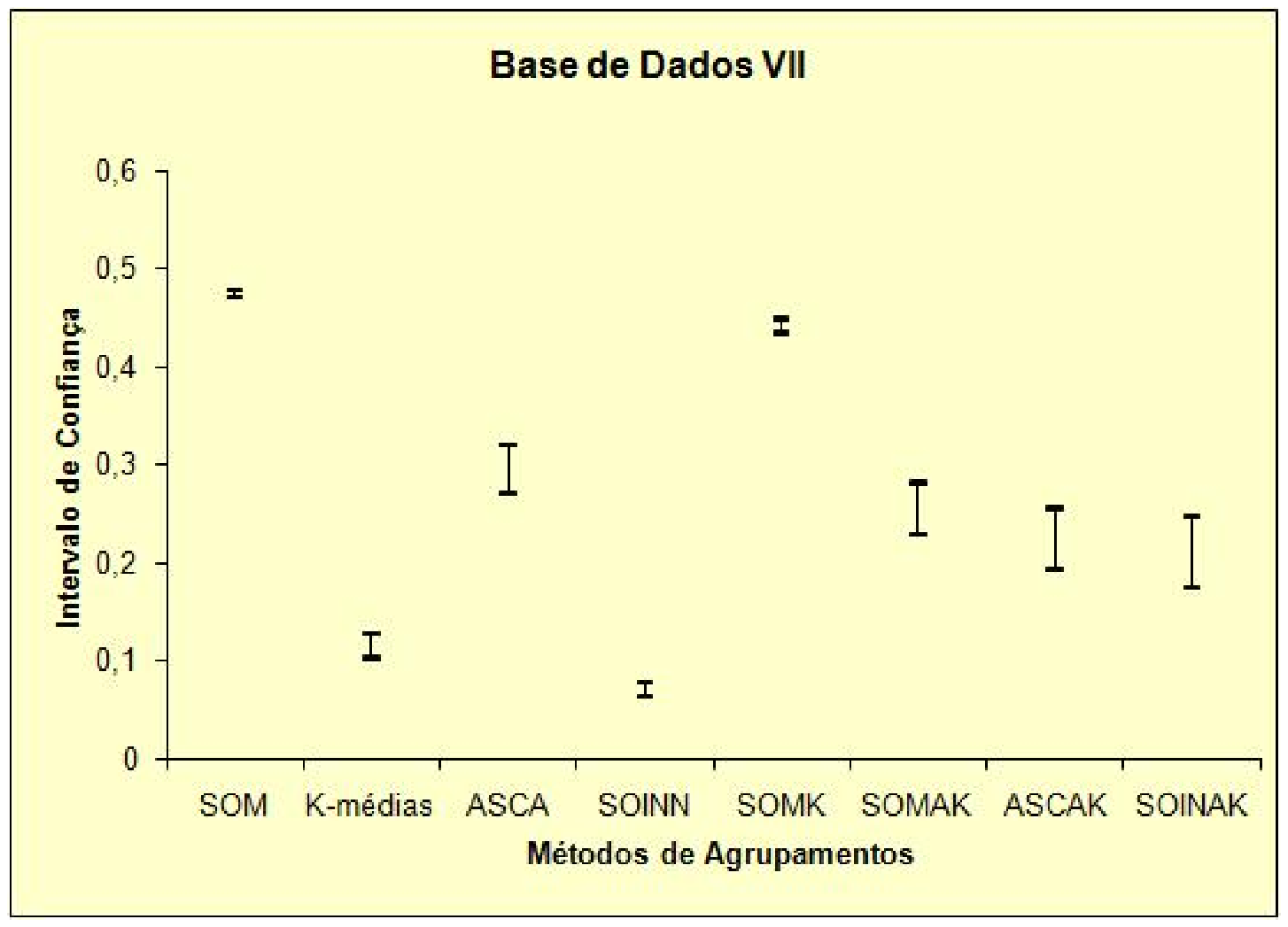,width=56mm}
}
\subfigure[]{
\epsfig{figure=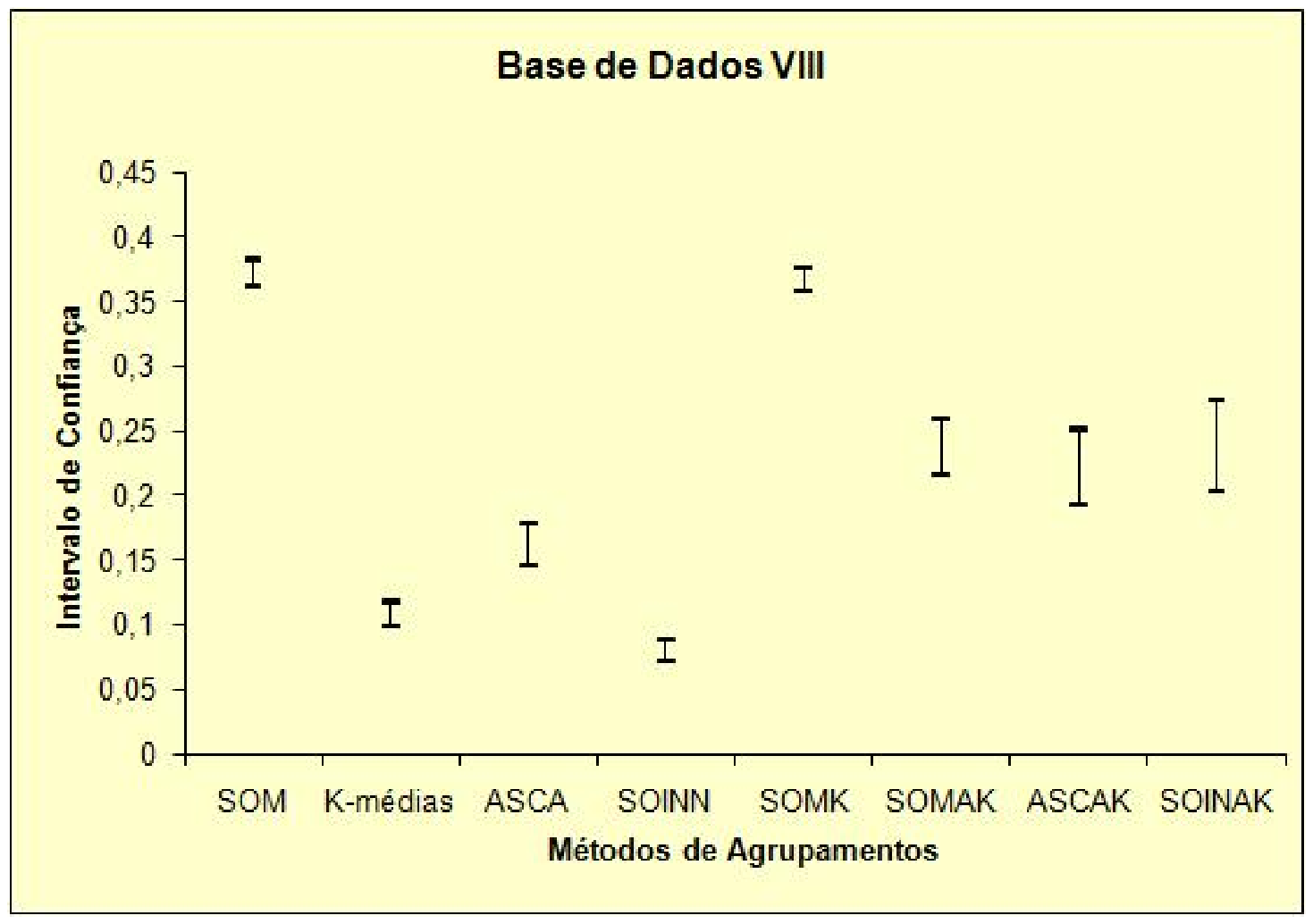,width=56mm}
}
\vspace{-14pt}
\caption{Intervalos de confiança da base \emph{CMC} (a) e \emph{Dermatology} (b).}
\label{fig:GraficoIntervaloConfiancaIV}
\end{figure*}

Para as bases \emph{Lines}, \emph{Highleyman}, \emph{Spherical} e \emph{Simple Problem} (Figuras \ref{fig:GraficoIntervaloConfiancaI} (a), \ref{fig:GraficoIntervaloConfiancaII} (a), \ref{fig:GraficoIntervaloConfiancaII} (b) e \ref{fig:GraficoIntervaloConfiancaIII} (a)), K-médias possui o menor valor médio e essa diferença é estatisticamente significativa. 
Já as bases \emph{Balance Scale}, CMC e \emph{Glass} (Figuras \ref{fig:GraficoIntervaloConfiancaIII} (b), \ref{fig:GraficoIntervaloConfiancaIV} (a) e \ref{fig:GraficoIntervaloConfiancaV} (b)), SOM possui o menor valor médio e essa diferença é estatisticamente significativa. Para as bases \emph{Banana} e \emph{Dermatology} (Figuras \ref{fig:GraficoIntervaloConfiancaI} (b) e \ref{fig:GraficoIntervaloConfiancaIV} (b)), SOINN possui o menor valor médio e essa diferença é estatisticamente significativa. Por fim, na base \emph{Diabetes} (Figura \ref{fig:GraficoIntervaloConfiancaV} (a)), SOMK possui o menor valor médio e essa diferença é estatisticamente significativa. As Figuras \ref{fig:GraficoIntervaloConfiancaI} a \ref{fig:GraficoIntervaloConfiancaV} mostram que K-médias possui o menor IC para 40\% dos dados seguido dos métodos SOM (30\%), SOINN (20\%) e SOMK (10\%) respectivamente. Assim, o K-médias possui a menor variabilidade e essa diferença é estatisticamente significativa.

Portanto, SOINAK apresentou um bom desempenho, e SOINN mostrou resultados significativos com um nível de confiança de 95\% para a maioria dos dados. Presume-se, que SOINAK é o melhor método de agrupamento avaliado neste trabalho.

\section{Conclusão}
\subsection{Resultados Obtidos}
O objetivo deste artigo foi propor os métodos de agrupamentos SOMAK, ASCAK e SOINAK, onde SOINAK se mostrou capaz encontrando um ótimo desempenho quando comparado com as técnicas de agrupamento em dois estágios. Os algoritmos de agrupamento foram testados para dados diretos e dados treinados pelos modelos auto-organizáveis (SOM e SOINN) e ASCA numa primeira etapa seguido de AK como segunda etapa dos métodos propostos.

SOINAK têm se mostrado melhor que SOMK, SOMAK e ASCAK por ter encontrado um melhor desempenho quando comparado com os métodos de agrupamento em dois estágios (Tabelas \ref{tab:ResultadosEntropiaDadosSinteticos} e \ref{tab:ResultadosEntropiaDadosReais}). Foram observados a medida de entropia e o tempo computacional, e os resultados experimentais confirmam com o uso do Teste \emph{t} com um nível de significância de 5\%.

\subsection{Limitações e Perspectivas Futuras}
Uma limitação de dois dos métodos propostos, ASCAK, SOINAK, é o alto custo computacional. Já o método SOMAK não tem tempo computacional alto. Dentre as perspectivas futuras está a possibilidade de verificar o tempo de outras abordagens híbridas de agrupamento vistas na literatura, objetivando hibridizar esta nova técnica de agrupamento com ASCAK e SOINAK, de modo a diminuir o custo computacional. Pretende-se usar outras técnicas de agrupamento na fase de pré-processamento dos padrões de entrada da rede neural, como método de e\-li\-mi\-na\-ção de ruído.

Outra possibilidade é ajustar os métodos ASCAK e SOINAK com a finalidade de reduzir o tempo computacional quando comparado com os métodos descritos neste trabalho. O primeiro método que será observado é o ABSOM \cite{nath:ShengChai2008}, que trabalha bem na análise de agrupamento em dois estágios, quando é usado como uma técnica de pré-processamento, além de apresentar resultados melhores quando comparado com a rede de Kohonen \cite{nath:Kohonen1998}. 

Com a utilização de SOINN combinada ao AK sendo um sistema híbrido de reconhecimento de padrões, verificamos a necessidade de automatizar os parâmetros de configuração da rede SOINN e do AK, devido à quantidade e à dificuldade de escolha dos parâmetros adequados. Assim, será testado a possibilidade de usar outros algoritmos de otimização global \cite{nath:Goldberg1989} e \cite{computacao-evolutiva-2002}, resultando o sistema híbrido de reconhecimento de padrões num funcionamento independente e sem a necessidade de ajuda do ser humano.

Podemos observar, que o benefício do SOMAK é a redução do tempo e de ruídos (em virtude de se usar um método baseado em dois estágios). Os desempenhos dos métodos SOMAK e SOMK foram comparados usando uma medida de entropia \cite{nath:Tan2006}. SOMAK apresentou valores menores (medida de entropia e tempo computacional) quando comparada ao método de agrupamento SOMK, para a maioria dos dados usados nestes artigos \cite{nath:Souza2009a} \cite{nath:Souza2009b}, mostrando ser melhor que SOMK.

Os resultados apresentados para os dados gerados pelo método Monte Carlo \cite{nath:Milligan1980} e \cite{nath:Milligan1985}, que tem como propósito gerar as bases de dados, mostraram que o sistema híbrido ASCAK neste trabalho é melhor do que os demais sistemas híbridos (SOMK e SOMAK), por ter encontrado um melhor desempenho quando comparado com essas duas técnicas.

Os resultados experimentais mostram um desempenho melhor do SOINAK comparado com SOMK, SOMAK e ASCAK; apresentando vantagens, como: redução de ruídos, quantidade menor de parâmetros para serem configurados, não necessita saber \emph{a priori} o número de agrupamentos como K-médias (ou AK) e informar os centróides iniciais. Portanto, SOINAK é um método de agrupamento robusto, podendo ser aplicado a diferentes tipos de problemas de agrupamento para obter resultados mais promissores.

\section*{Agradecimentos}
Os autores agradecem ao CNPq, CAPES, e FACEPE (agências brasileiras de fomento à pesquisa) pela ajuda financeira.

\renewcommand\refname{REFER\^{E}NCIAS}

\bibliographystyle{lnlm}
\bibliography{lnlm}
\end{document}